\documentclass{bmcart}

\usepackage[utf8]{inputenc} 
\usepackage{graphicx}

\usepackage{tabularx}

\usepackage{array}
\usepackage{ragged2e}
\newcolumntype{P}[1]{>{\RaggedRight\hspace{0pt}}p{#1}}

\newcommand{\fixmetext}[1]{{\color{black}#1}}

\newcommand{\fixmetextt}[1]{{\color{black}#1}}

\startlocaldefs
\endlocaldefs

\begin{document}

\begin{frontmatter}

\begin{fmbox}
\dochead{Research}


\title{Using semantic clustering to support situation awareness on Twitter: The case of World Views}


\author[
addressref={aff1},                   
]{\inits{C}\fnm{Charlie} \snm{Kingston}}
\author[
addressref={aff2,aff1},
corref={aff2}, 
email={j.r.c.nurse@kent.ac.uk}
]{\inits{JRC}\fnm{Jason R. C.} \snm{Nurse}}
\author[
addressref={aff1},
email={ioannis.agrafiotis@cs.ox.ac.uk}
]{\inits{I}\fnm{Ioannis} \snm{Agrafiotis}}
\author[
addressref={aff3},
email={}
]{\inits{I}\fnm{Andrew} \snm{Burke Milich}}


\address[id=aff1]{%
	\orgname{Department of Computer Science, 
		University of Oxford},
	\city{Oxford},
	\cny{United Kingdom}
}
\address[id=aff2]{%
	\orgname{School of Computing,
		\newline University of Kent},
	\city{Canterbury},
	\cny{United Kingdom}
}
\address[id=aff3]{%
	\orgname{Stanford University},
	\city{Stanford},
	\cny{USA}
}


\begin{artnotes}
\end{artnotes}

\end{fmbox}


\begin{abstractbox}

\begin{abstract} 
In recent years, situation awareness has been recognised as a 
critical part of effective decision making, in particular for 
crisis management. One way to extract value and allow for better 
situation awareness is to develop a system capable of analysing a 
dataset of multiple posts, and clustering consistent posts into 
different views or stories (or, `world views'). However, this can 
be challenging as it requires an \textit{understanding} of the data, 
including determining what is consistent data, and what data 
corroborates other data. Attempting to address these problems, 
this article proposes \textit{Subject-Verb-Object Semantic Suffix 
	Tree Clustering} (SVOSSTC) and a system to support it, with a
special focus on Twitter content. The 
novelty and value of SVOSSTC is its emphasis on utilising the 
Subject-Verb-Object (SVO) typology in order to construct 
semantically consistent world views, in which individuals---particularly 
those involved in crisis response---might achieve 
an enhanced picture of a situation from social media data. To 
evaluate our system and its ability to provide enhanced situation 
awareness, we tested it against existing approaches, including 
human data analysis, using a 
variety of real-world scenarios. The results indicated a 
noteworthy degree of evidence (e.g., in cluster granularity 
and meaningfulness) to affirm the suitability and rigour of 
our approach. Moreover, these results highlight this article's 
proposals as innovative and practical system contributions to 
the research field. 
\end{abstract}




\end{abstractbox}
%

\end{frontmatter}




\section{Introduction}\label{sec:introduction}
The quantity of information available online is astounding.
Social media has played a key part in this boom of content, as it has emerged as 
a central platform for communication and information sharing, allowing users to post 
messages related to any event or topic of interest~\cite{karandikar2010clustering}.
Possibly one of the most significant uses of social media today is its ability to 
help understand on-going situations. Situation awareness has been recognised as a 
critical part of effective decision making, in particular for crisis management 
scenarios~\cite{blandford2004situation}. Twitter, for instance, is regularly used 
as a news breaking mechanism to provide near-real-time observations about 
situations~\cite{kwak2010twitter}. By leveraging the public's collective intelligence, 
emergency responders may be able create a holistic view of a situation, allowing 
them to make the most informed decisions possible. 

There are two key problems that prevent users from gathering valuable and actionable 
intelligence from social media data. The first is the massive amount of information
shared leading to information overload, and the second is the proliferation of mistaken 
and inadvertent misinformation~\cite{rodriguez2014quantifying}. Taking the 2013 Boston 
bombing as an example, the number of tweets posted reached 44,000 per minute just 
moments after the attack~\cite{withnall2013twitter}. This was simply too much data to
consume, even for official services. To exacerbate the problem, many post-mortem reports 
indicated that much of this information was inaccurate, with on average, only 20\% 
presenting accurate pieces of factual information~\cite{smithss2013al}. 

In order to combat the emerging phenomenon of information overload and to support 
better understanding of situations using large amounts of data, there is a growing 
need to provide systems and tools that can analyse data and provide enhanced insight.
One of the approaches that has been suggested is that of creating `world views' to allow 
better understanding of a situation. A world view is a cluster of consistent messages that 
gives a possible view of a scenario~\cite{rahman2014data}. It contains key aspects in 
support of an individual's perception of environmental elements with respect to time or 
space; the comprehension of their meaning; and the projection of their status after some 
variable has changed, such as a predetermined event~\cite{endsley1995toward}. By presenting 
users with a more complete, consistent, and corroborative picture of a situation, the 
notion of world views can help to enhance a user's awareness in a scenario or situation. We believe that this enhanced awareness can serve as a crucial starting point (i.e., not the full solution) to address the problem of misinformation in social media.

In this paper, we aim to address some of these issues through the development and evaluation 
of a system supporting user and organisational situation awareness using social media data. 
The goal of our system is to facilitate the analysis of datasets of multiple posts, and 
allow the clustering of consistent posts into different world views. These views can provide
valuable insight into on-going scenarios (e.g., crises), that could then lead to better 
decision-making (e.g., where to send emergency responders as in the case of the London 
Riots~\cite{guardian2012riot}). A main research challenge that we seek 
to tackle here is the creation of a novel system that can \textit{understand} data through 
the application of Natural Language Processing (NLP) techniques, and determine consistent 
and corroborated information items. This work is intended to complement our existing 
efforts of building a suite of tools for individuals and organisations that can allow
actionable intelligence to be gained from open source 
information~\cite{nurse2012using,nurse2013building,nurse2014two,nurse2015tag}. These tools
would be tailored for analysis of online content while also possessing interfaces suited
for human cognition.

The remainder of this paper is structured as follows. Section~\ref{section:litrev} 
reflects on the relevant literature in the fields of social media, misinformation, 
and situation awareness. Next, in Section~\ref{section:sysapp} we present our world 
view extraction approach to understanding social media data. Here we also detail the 
use, application, and scope of the system. We provide an overview of the system 
architecture in Section~\ref{section:archi}. In Section~\ref{section:evadis}, we 
report on an evaluation of our system involving a comparison to a number of existing 
systems that have similar aims. Finally, we conclude the article in 
Section~\ref{section:con} and consider future work.

\section{Related Work}
\label{section:litrev}
The proliferation of social media has made it a practicable medium to acquire 
insights about events and their development in environments~\cite{scott2012social}. The role of social media in 
natural disaster crisis management became clear during the 2010 Haiti 
earthquake~\cite{keim2010emergent}, and has increased significantly since 
then~\cite{yin2012using}. The research community has focused on two general areas
to gain the most value from social media especially with such situations in mind. 
These include, tackling misinformation and its spread, and broad approaches to 
understanding situations.

\subsection{The Misinformation Problem}

Misinformation can easily spread in a network of people, highlighting the importance 
of designing systems that allow users to detect false information. In traditional 
communication media, machine learning and Natural Language Processing (NLP) are 
often used to automate the process. However, social networking services like 
Twitter suffer from intrinsically noisy data that embodies language use that 
is different from conventional documents~\cite{yin2012using}. On top of this, 
Twitter restricts the length of the content published, limiting the usefulness 
of traditional trust factors (such as length of content) as an indicator of 
information quality~\cite{castillo2011information}. This results in the accuracy 
of automated methods with social media data being highly limited, and so manual 
intervention is often required.

One approach taken by Procter et al.~\cite{procter2013reading} to understand 
widespread information on Twitter used manual content analysis. The approach 
utilised code frames for retweet content in order to categorise information 
flows (e.g., a report of an event), using the groupings to explore how people 
were using Twitter in the corresponding context. By using a variety of tweet 
codes, categories such as media reports, rumours (misinformation), and reactions 
were identified. In particular, rumours were identified as tweets where users 
had published content, without providing a reference (e.g., an external link).

Other approaches have also attempted to use assessments of individual information
items (e.g., tweets) using trust metrics~\cite{nurse2014two,cho2015survey,nurse2013communicating}. 
While these provide 
a rigorous and automated approach, they often require reliable and a good quantity
of metadata to make appropriate judgements. Additional attempts at addressing 
misinformation issues have also sought to train machine learning classifiers. 
These would assist with the identification of rumours and low-quality 
information~\cite{castillo2011information}. Again however, 
these often require the manual annotation of misinformation to help with 
classification tasks.

\subsection{Approaches to Understanding Situations}

Some of the earliest work which attempted to understand situations from social 
media data was by Sakaki et al.~\cite{sakaki2010earthquake} who manually defined 
a set of keywords relevant for the types of events they wanted to detect (earthquake, 
shaking, and typhoon). They used a Support Vector Machine (SVM) to classify each 
tweet based on whether it referred to a relevant event (i.e., an event described by any of the keywords)) or not. This approach was limited as the set of keywords needed to be defined manually 
for each event, and hence a separate classifier needed to be trained. 

By acknowledging the importance of syntax and semantics, the approach taken 
by Vosoughi~\cite{vosoughi2015automatic} used a Twitter speech-act classifier which 
exploited the syntactic and semantic features of tweets. The classifier was successful 
in analysing assertions, expressions, questions, recommendations, and requests, but 
it only used a single tweet as the basis for classification. This skewed the cluster 
towards the initial seed tweet, impacting the overall awareness gained by a 
situation. However, Vosoughi's research was very successful in predicting the 
veracity of rumours by analysing (i) the linguistic style of tweets; (ii) the 
characteristics of individuals spreading information; and (iii) the network 
propagation dynamics. The veracity of these features extracted from a collection 
of tweets was then generated using Hidden Markov Models (HMMs)~\cite{vosoughi2015automatic}.

Yin et al. propose a system that uses a variety of techniques 
to cluster tweets in an effective way~\cite{yin2012using}. It works by initially 
gathering data in the data capture component, processed using burst detection, text 
classification, online clustering, and geo-tagging, and then visualisation by the 
system user. A crucial aspect of the methodology used by Yin et al. is the 
clustering component, which motivates our research. Specifically, no prior 
knowledge of the number of clusters is assumed. Within the domain of crisis 
management, this is especially important as crises evolve over time. This means 
that partition clustering algorithms such as $k$-means, 
along with hierarchical clustering algorithms (which require a complete 
similarity matrix) are not appropriate in this system. Instead, the system extends 
Group Average Clustering (GAC), an agglomerative algorithm which 
maximises the average similarity between document pairs~\cite{yang1998study}.

In a similar vein, Suffix Tree Clustering (STC)~\cite{ukkonen1995line} 
has been used in Web search engines to group 
information~\cite{ilic2014suffix}. In particular, the suffix tree 
algorithm is well-suited to domains where the number of words in each document is 
very small~\cite{branson2002clustering}, such as in 
social media platforms as Twitter. 

\fixmetext{Janruang et al.~\cite{janruang2011semantic}, expand STC approaches 
by proposing an algorithm named Semantic Suffix Tree Clustering (SSTC) which utilises a Subject Verb Object (SVO) classification
to generate more informed names in clustersfor web search results. The main difference and novelty of our approach is the use of Ukkonen’s online, linear time, and space-efficient algorithm~\cite{ukkonen1995line}. SSTC uses a less space efficient algorithm to ensure all phrases fully appear in the suffix tree~\cite{janruang2011semantic}. Furthermore, unlike SSTC, our approach will add a trustworthiness assessment to the final clustering process. As SVOSSTC was designed to extract world views from social media information, understanding the reliability and dissemination of information is critical. Thus, SSTC and SVOSSTC both use SVO classification and WordNet as a similarity metric, but SVOSSTC uses a different semantic suffix tree construction algorithm and assesses clusters' trustworthiness. The applications of STC 
to create world views from social media data have yet to be fully explored, since SSTC was used for web searches, and 
they serve as a point of interest for our research.}

The notion of world views has not been analysed at large by academics. Despite 
this, one successful approach for using world views to better understand situations 
was proposed by Rahman~\cite{rahman2014data}, who used a 
number of Tactical Situation Objects (TSOs) to create a set of internally 
consistent clusters (world views), ranked by an initial provenance metric. TSOs 
represent an intermediate form of a tweet, allowing natural language to be 
encoded into a structured XML for the system to process. However, the use of 
TSOs encoded as XML objects in order to structure tweets is a very
simplified assumption in an unstructured real-world environment 
such as Twitter. Other recent literature has also identified the difficulties
of understanding, summarising or assessing this problem~\cite{rudrapal2018auto,kwona2013opin}

\section{System Approach}
\label{section:sysapp}

\begin{figure}[ht!]
	\centering
	\includegraphics[width=.88\textwidth]{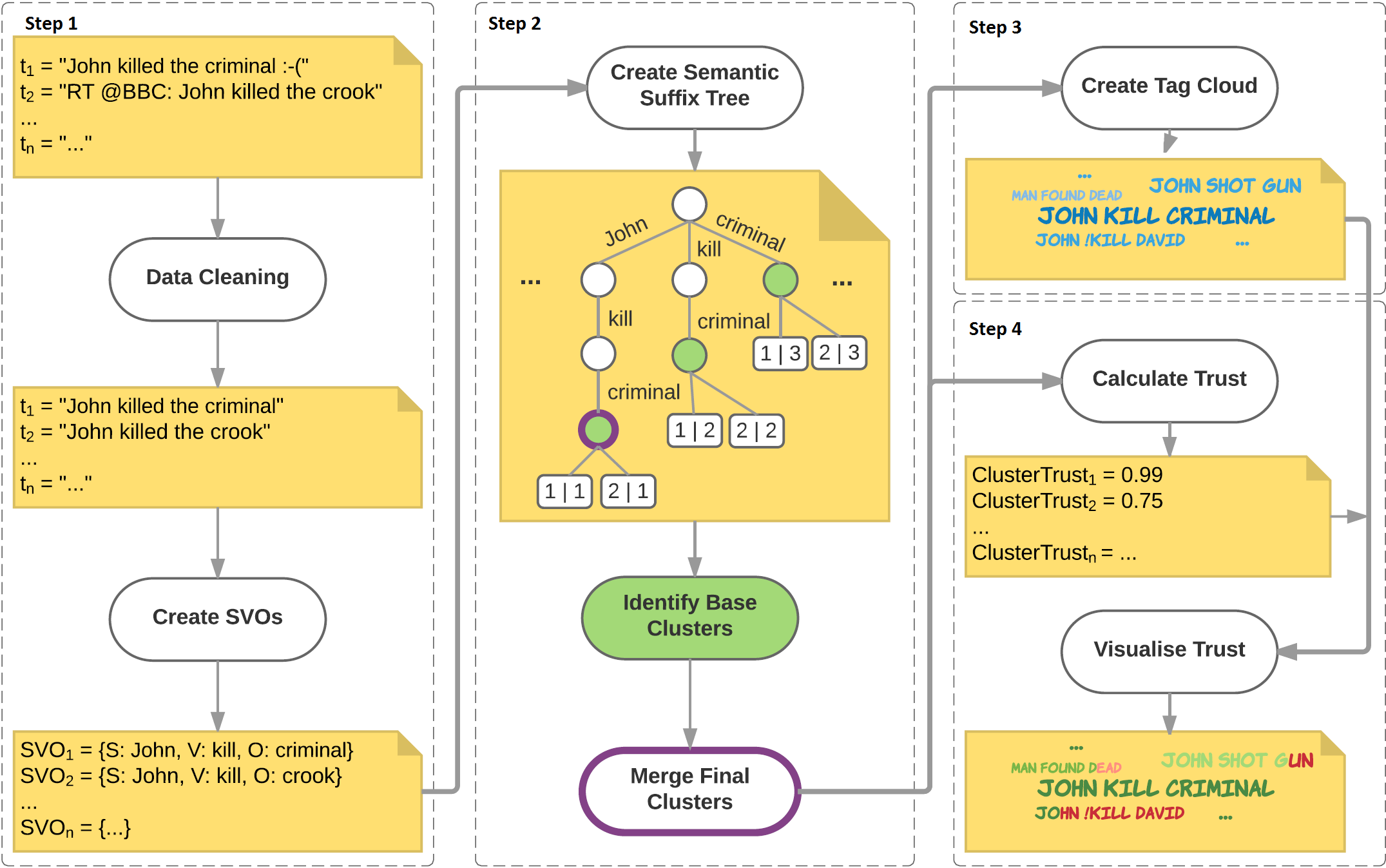} 
	\caption[Overview of the approach to creating our system]{Overview of 
		the approach to creating our system}
	\label{fig:approach-flow-chart}
\end{figure}

The goal of our system is to support users in the analysis of multiple posts, 
and allow the clustering of consistent posts into different views of a scenario
(i.e., world view). The approach that we have taken to fulfil this aim consists
of four main steps, which are shown in Figure~\ref{fig:approach-flow-chart}.

In what follows, we present these steps in further detail and explain how
they work towards addressing our aim.

\subsection{Standardising Data with Subject-Verb-Object Tuples}
\label{section:step1}

Data standardisation is challenging with social media, in particular on 
Twitter, because (i) tweets are noisy and ambiguous; (ii) there is no 
well-defined schema for the various events reported via Twitter; and 
(iii) it is not trivial to extract information from unstructured text. 
We believe that an approach using the 
Subject-Verb-Object (SVO) typology could be a viable solution, 
as it is often considered to be the dominant sentence structure in social media communication~\cite{schrading2015whyistayed}.

The SVO representation~\cite{tomlin2014basic} is a linguistic structure 
where the subject comes first, the verb second, and the object third. 
SVO languages, such as English, almost always place relative clauses 
after the nouns they modify and adverbial subordinators before the 
clause modified. An example of the SVO linguistic structure applied to 
a basic phrase (\textbf{P1}) is shown below:

\textbf{P1}: \textit{``David ate lunch''} $\Rightarrow \{(S\colon \textit{David}, \ V\colon \textit{ate}, \ O\colon \textit{lunch})\}$

\noindent
As discussed in Section~\ref{section:litrev}, there have been many 
attempts to extract value from social media data by using an entire 
tweet as the input corpus to a system. However, the noise, structure, 
and complexity of social media data often mean that it is not well 
suited to the task of information extraction. The SVO representation 
is useful as it alleviates the issue of data standardisation by 
producing structured tuples of linguistic information. By using the 
SVO representation in our approach, we aim to partially 
address the issue of unstructured data in social media. Below we 
present a real tweet (\textbf{T1}) and show its 
SVO representation:

\textbf{T1}: \textit{``RT @ABC: The FAA issued a flight restriction''} \\ \indent \indent $\Rightarrow \{(S\colon \textit{FAA}, \ V\colon \textit{isssued}, \ O\colon \textit{restriction})\}$

In our approach, we apply a series of data cleaning functions to 
preprocess the tweet before converting it to an SVO representation. 
These functions include (i) syntax cleaning to reduce inflectional 
and derivationally related forms of a word to a common base form; 
(ii) tweet cleaning to remove hashtags, retweets, and other 
discourse found on Twitter; and (iii) slang lookup to identify 
unfamiliar words and convert them to standard dictionary English. Each tweet 
is then parsed into $n$ SVO tuples, along with an identifier 
corresponding to the original tweet. For example, in \textbf{T2} 
shown below, we 
have two SVO tuples corresponding to each of the possible SVO 
representations of the tweet: 

\textbf{T2}: \textit{``New images show suspect: Massachusetts police released several images''} \\ \indent \indent $\Rightarrow \{(S\colon \textit{images}, \ V\colon \textit{show}, \ O\colon \textit{suspect}), (S\colon \textit{police}, \  V\colon \textit{released}, \ O\colon \textit{images})\}$

As each tweet in our system is an ordered 
tuple of length three, as opposed to a long sequence of unordered 
keywords, our clustering methodology is able to produce more succinct 
and relevant cluster labels.

We have also considered how lexical databases such as 
WordNet~\cite{miller1995wordnet} can enhance our understanding of 
each of the components in the SVO representation. WordNet contains 
groups of synonyms, called synsets, which record the relationship 
between the members of the set. By identifying the synsets from each 
component of the SVO representation, our approach exploits semantic 
similarity in order to reduce the future overlap of semantically 
equal (but syntactically different) cluster labels. Hence, this 
approach allows us to produce semantically consistent clusters.

The novelty of our approach lies in the analysis of how effective 
the SVO representation can be in structuring language in order 
to extract valuable meaning from data. While there is always a potential
that data might be lost in any automated information extraction approach, we note that a significant 
proportion of tweets typically fit into the SVO 
structure~\cite{schrading2015whyistayed}. Consequently, we are 
therefore motivated by the potential usage of this technique in creating concise and meaningful 
descriptions for different world views. It is these world views that
we aim to utilise to increase situation awareness.

\subsection{Applying Suffix Tree Clustering to Social Media Data}
\label{section:step2}

A suffix tree is a compressed trie (also known as a digital tree) 
containing all possible suffixes of a given text as the keys, and 
their position in the text as the values~\cite{lobhardXXXXsuffix}. 
Suffix trees are particularly useful as their construction for a 
string $S$ takes time and space linear in the length of $S$. We 
define, for a string $S \in \Sigma *$ and $i,j \in \{1, ..., |S|\}, 
i \le j$, the substring of $S$ from position $i$ to $j$
as $S[i,j]$, and the single character at position $i$ as $S[i]$. 
The suffix tree for the string $S$ of length $n$ is 
defined as a rooted directed tree with edges that are labelled 
with nonempty strings and exactly $m$ leaves labelled with 
integers from $1$ to $j$. This is such that: (i) each internal 
node other than root has at least two children; (ii) no two edges 
out of one node have edge labels beginning with the same character; and
(iii) for any leaf $i$, the concatenation of the path labels from 
root to leaf $i$ is $S[i,m]$.

Since such a tree does not exist for all strings $S$, a terminal 
symbol that is not seen in the string (usually denoted 
\textit{``$\$id$''}) is appended to the string. This ensures that 
no suffix is a prefix of another, and that there will be $n$ leaf 
nodes, one for each of the $n$ suffixes of $S$. A slight variation 
of the suffix tree, used in our approach, is a Generalised Suffix 
Tree (GST)~\cite{lobhardXXXXsuffix}. A GST is constructed for a 
set of words instead of single characters, which makes it much 
more effective for the purposes of understanding sentences, and 
hence situations. Each node of the GST represents a group of documents 
and a phrase that is common to all of them. The best algorithm for 
suffix tree construction is Ukkonen's algorithm~\cite{ukkonen1995line} 
which is linear time for constant-size alphabets. In our approach, 
we adapt Ukkonen's algorithm in order to apply it our GST representation.

Once constructed, the suffix tree approach allows several operations 
to be performed quickly. For instance, locating a substring in $S$, 
locating matches for a regular expression pattern, and many more. 
However, suffix trees can also be utilised for the purposes of 
clustering. In particular, Suffix Tree Clustering (STC) 
has been widely used to enhance Web search results, where short 
text summaries (also known as `snippets') are 
clustered~\cite{zamir1998web,branson2002clustering}. 
The STC algorithm groups input documents according to the phrases 
they share, on the assumption that phrases, rather than keywords, 
have a far greater descriptive power due to their ability to retain 
relationships and proximity between words. We aim to exploit this 
in our approach in order to produce highly descriptive cluster labels.

The clustering methodology has two main phases: base cluster 
discovery and base cluster merging. In the first phase, each of 
the documents (tweets) are built into the GST, where each internal 
node forms an initial base cluster. The second phase then builds 
a graph representing the relationships between the initial base 
clusters using a similarity measure. This measure is defined as 
the similarity in document sets. Effectively, this criterion 
requires that each cluster must have the most specific label 
possible to avoid unnecessary, less descriptive, but semantically 
identical clusters. The clusters are then merged into final 
clusters if they satisfy the similarity measure. An example of a 
STC implementation for Web search results is 
Carrot2~\cite{stefanowski2003carrot2}.

An advantage of STC is that phrases are used both to discover 
and to describe the resulting groups, achieving concise and 
meaningful descriptions. Furthermore, methods that utilise 
frequency distribution often produce an unorganised set of keywords. 
We overcome this issue by using STC in our approach. However, 
some previous attempts at using STC have been limited. For 
example, if a document does not include any of the exact phrases 
found in other documents then it will not be included in the 
resulting cluster, even though it may be semantically identical. 
Acknowledging this, our approach combines semantic reasoning 
through utilising WordNet synonym rings (synsets) as part of 
our SVO tuples (tweets), as mentioned in Section~\ref{section:step1}, 
in order to alleviate the issue of semantic consistency. 
Therefore, our approach introduces \textit{Subject-Verb-Object 
	Semantic Suffix Tree Clustering} (SVOSSTC).

To enable our SVOSSTC approach, we relax the suffix tree 
constraint that each internal node other than root has at least 
two children, and enforce the constraint that each label must 
only be a single word. Using this, we build upon the successful 
use of STC in Web search results, by exploiting the distinct 
similarity between snippets and tweets, in order to trial the 
effectiveness of STC for social media data. Combining the SVO 
approach outlined in the section above with Semantic 
Suffix Tree Clustering (SSTC), we propose the following 
algorithm in our wider approach:

\begin{enumerate}
	\item Generate SVO representation for each tweet (as 
	previously defined);
	\item Create a \textit{Subject-Verb-Object Semantic Suffix 
		Tree} (SVOSST) $T$ with a single root node;
	\item For each SVO, ascertain the associated WordNet synset 
	for each part of the SVO representation;
	\item For each word in the SVO representation, if the overlap 
	between the synset of the current word and the synset at node 
	$n$ is $\ge 1$, then we create a link in $T$, else we add the 
	synset of the current word to $T$;
	\item After each SVO has been added, we insert a label which 
	includes the tweet identifier (terminal symbol) and the 
	starting branch for the feature word to $T$;
	\item Let each subtree $T_0, T_1, ..., T_i$ be a concept 
	cluster, and each node a cluster that has a set of documents 
	to be a member;
	\item Each base cluster is then formed using a post-order 
	traversal of the nodes along with the corresponding label;
	\item Merge base clusters $c_i, c_j$ with $|c_i \cap c_j| = |c_a|$, 
	then delete $c_a$, or $|c_i \cap c_j| = |c_b|$, then delete $c_b$; and
	\item Output final clusters.
\end{enumerate}

The novelty and contribution of our work is the definition of an approach 
which draws on, and extends, existing work from other fields. It applies 
SVOSSTC to social media data, exploiting the STC algorithm's low complexity 
and successful 
applications in Web search results. In particular, our contribution improves 
upon the basic STC algorithm by using semantic information provided by 
lexical resources such as WordNet. This increases the likelihood 
that semantically consistent clusters will be created. Due to the 
SVO structure utilised in our approach, the scope of semantic 
similarity is narrowed, ensuring that the cluster labels are as 
descriptive as possible. For example, traditional approaches to 
analyse semantic similarity in STC do not consider the structural 
formation of the sentence, and hence increase both the complexity 
of the algorithm and obfuscation of the output by incorrectly 
considering the semantics of structurally different 
words~\cite{dang2013wordnet}. The combination of 
the SVO representation and STC is crucial to our approach, as it 
produces non-contradictory clusters with a precise semantic meaning.
This, as will be discussed below, offers several advances for users in
gaining a better situational awareness, especially in situations of crisis.

\subsection{Creating World Views with Tag Clouds}

A tag cloud, or word cloud, is a visualisation technique for textual 
data, where size, colour, and positioning are used to indicate 
characteristics (such as frequency and prominence) of the 
words. Tag clouds are often used to display 
summaries of large amounts of text, especially with regard to providing 
a quick perspective of a situation~\cite{nurse2015tag}. In our approach, we 
use tag clouds as the primary visualisation mechanism to display world views
as output from our SVOSSTC approach. We follow key visual design principles
for situation awareness~\cite{lanfranchi2014visual} to increase their usefulness.

Specifically, we aim to (i) use a filtering mechanism to alter the level 
of granularity when clustering; (ii) utilise standardised vocabularies 
with the SVO representation; (iii) highlight the correlation between 
elements with STC; and (iv) provide a flexible pathway for exploring 
related information. The approach taken allows users to 
see both the SVO representation for highlighting general trends, as well 
as an in-depth overview of all the available tweets that create each 
cluster. One key difference is that the words in our tag cloud are not 
independent from each other, but are considered as tags of ordered 
tuples. Therefore, the data is contextualised into a wider picture, which is an important feature for enhancing situation awareness. 

\subsection{Analysing Trust in World Views}

In order to address the problem of 
misinformation in social media data, as discussed in Section~\ref{section:litrev}, we would need to consider how the world views can be used in the 
context of trustworthiness. A potential solution to the problem of 
misinformation was proposed in our earlier work~\cite{nurse2015information}, 
by introducing the notion of information-trustworthiness measures. There,
we acknowledge world views as a valuable mechanism in a wider system that can 
analyse the trustworthiness of information.

Trustworthiness can be considered as an extension of quality (and hence, value) 
in social media data~\cite{kelton2008trust}. It is also perceived that 
trustworthiness is the likelihood that a piece of information will preserve a
user's trust in it~\cite{nurse2015information}. Therefore, our work also uses 
quality and trust metrics to assess the social media content, and then, based 
on the values attained and world views produced, informs users of the 
trustworthiness of the content.

\begin{figure}[ht!]
	\centering
	\includegraphics[width=0.9\textwidth]{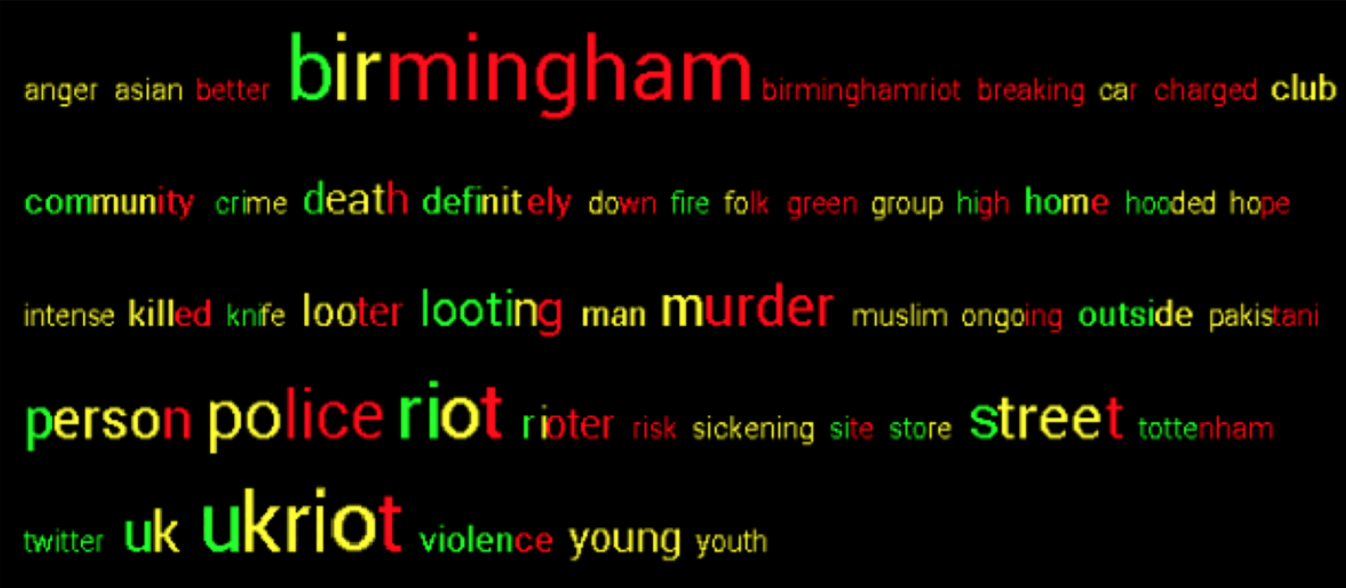} 
	\caption[Tag cloud generated from the 2011 London riots]{Tag cloud generated 
		from the 2011 London riots~\cite{nurse2015tag}}
	\label{fig:tag-cloud-twist}
\end{figure}

Figure~\ref{fig:tag-cloud-twist} shows how we used coloured tag clouds of 
the 2011 UK Riots Twitter dataset to convey trustworthiness. Here, letters within
words were coloured according to the trustworthiness of the contexts in which 
they appear, with green being the most trustworthy and red the least. From our experimentation with this visualisation approach,
we were able to show that it is useful in decision-making, particularly in 
facilitating a quick, helpful, and accurate overview of a situation~\cite{nurse2015tag}.

We incorporate the notion of trust into our current work on world views by 
first creating a world view, then assessing how trustworthy each world view is based 
on trustworthiness assessment approaches. There are a variety of different trust 
factors that we considered in our approach, for instance, competence of source, 
location, user relationships, corroboration, timeliness, 
and popularity~\cite{nurse2011information}. In the context of our work, we focus 
on corroboration as a key trust metric from the perspective of world views.

Corroboration is a measure of how many different sources agree with the content of 
the information provided. Traditionally, clustering engines are 
limited in respect to corroboration for a number of reasons. Most notably, they 
can ignore negation in words, resulting in semantically inconsistent clusters. 
Other approaches cluster a range of tweets and take the size of 
the cluster as the sole representation of corroboration. This approach should
be treated carefully because users may be retweeting 
misinformation, which falsely identifies high corroboration amongst sources. 

Therefore, our approach analyses corroboration, and generally the trustworthiness
of a world view by considering three main factors. These are: (i) how much information is 
corroborated i.e., how large the clusters are (excluding retweets); (ii) identifying the
extent to which there are trusted (predefined) parties that can increase the cluster's 
trustworthiness if present; and (iii) looking to further corroborate clusters with 
external entities (e.g., news reports on websites) to automatically see which 
clusters may be more trustworthy. As a start, instead of using real-time identification of related entities, we have decided to use a predefined list of news agencies present on Twitter that we assume (for this case) are trustworthy, and then use
these in (iii) to corroborate the trustworthiness within each cluster.
We acknowledge that news
sources are not always correct or timely, therefore only use this as an extension
to our system that may add value. Examples of agencies we have currently included 
are ABC, BBC News, CBS News, CNN and Reuters. We postulate that a larger 
proportion of news agencies within a cluster results in a large corroboration, and 
hence potentially, a higher level of trust. Users of our system would be free
to include their own list of pre-defined trusted parties. We use the following 
algorithm in our approach:

\begin{enumerate}
	\item Generate world views using the algorithm that we have proposed earlier;
	\item Let $\textit{clusters} = $ set of all world views, and $n_i = |\textit{cluster}_i|$ (without retweets);
	\item Let $t = |$trusted news agencies$|$;
	\item For each $\textit{cluster}_i \in \textit{clusters}$, where $i = 1...y$ and $y = |\textit{clusters}|$, sort the clusters from largest to smallest. Then, $x_i = |\textit{cluster}_i| / $max$(n_i)$, where $cluster_i$ $ \in$ $clusters$;
	\item For each $\textit{tweet}_k$ in $\textit{cluster}_i$, observe if it is from a news agency as defined, and let $t_i = |$trusted news agencies in$ \ \textit{cluster}_i|$. Then, $c_i = t_i / t$;
	\item After each $x_i$ and $c_i$ has been established, calculate $s_i = (0.5 \times x_i) + (0.5 \times c_i)$, to give an output between 0 and 1; and
	\item Output $s_i, \forall i \in \textit{clusters}$.
\end{enumerate}

As can be seen above, we assign equal weighting (0.5 each) to the relative size of the cluster ($x_i$) and the relative number of news agencies in the clusters ($c_i$). While this value is somewhat arbitrary, our motivation for these weights is due to both being viewed as equally important factors in measuring corroboration. This is the sole instance of weights usage within our method. In future work, we could seek to examine what weights may be the most appropriate for use, or even opening weighting as an option for users to configure.

It is important to note that our approach is only one of many ways that the
problem of misinformation can be addressed. For instance, we could apply the 
trust metrics to each tweet (as done in~\cite{nurse2013supporting,gupta2014tweetcred}) 
in each cluster and combine these to produce a trustworthy 
rating per world view. Or, we may look to use machine learning approaches to
determine the veracity and credibility of the information (as intended 
in~\cite{castillo2011information,giasemidis2016determining}). 
The contribution and scope of our work derives instead from extracting, 
creating, and visualising world views. Our approach explores a new domain by focusing 
on social media content, rather than the existing encoding formats that are widely 
used in crises. To take this further, we would blend a range of open-source and 
closed-source intelligence in order to create a more complete picture for public 
users of our system or official situation responders.

\section{System Implementation}
\label{section:archi}

The system architecture is split in three main components, each implemented in  
Python~\cite{python2016python}. These components are: \texttt{DataCleaning}, 
\texttt{Clustering}, and \texttt{Visualisation}. Each component is an identifiable 
part of the larger program, and provides a discrete group of related functions. By 
developing the system in a modular way, we were able to define clear interfaces 
which are crucial for the extensibility of the system in the future.

\subsection{Data Cleaning}

The \texttt{DataCleaning} component uploads, processes, and cleans social media data 
using a variety of pragmatic techniques that we have developed, combined with 
packages from the Natural Language Toolkit~\cite{nltk2016nltk} (i.e., libraries 
and programs for symbolic and statistical NLP for the English language).
The three main modules that we created to assist in this task are: the
\texttt{SyntaxCleaner}, which performs activities such as escaping HTML characters, 
removing unnecessary punctuation, and decoding the data to the ASCII format; the 
\texttt{TweetCleaner}, that removes URLs, emoticons, and Twitter-specific discourse 
such as mentions and hashtags characters; and the \texttt{SlangLookup} which is 
responsible for converting colloquial abbreviations such as \textit{``how'd''} and \textit{``m8''} to formal English (how did and mate). 

The \texttt{DataCleaning} component ensures that a tweet is translated to a (clean) natural language representation 
before the \texttt{SVO} function begins processing. The original tweets are 
also stored along with pointers from the beginning of each word in a clean 
tweet to its position in the original tweet. This enables the system to display 
both versions of the text to enhance readability and understanding. In the next phase, 
the system processes the clean tweet to obtain the SVO representations. 

Once the system has obtained an SVO representation, it uses the \texttt{Lemmatiser} 
and \texttt{VerbPresent} functions to further standardise the output for future 
clustering operations. In the \texttt{Lemmatiser} function, each component of the 
SVO tuple is transformed with a light stemming algorithm which utilises the 
\texttt{WordNetLemmatizer}~\cite{nltk2016nltk} function defined in the \texttt{NLTK} 
module. This function reduces inflectional endings to their base or dictionary 
form using the word itself and the corresponding part-of-speech tag (from 
\texttt{spaCy}~\cite{spacy2016spacy}) to establish the correct context of the word.

The \texttt{VerbPresent} function operates on the verb within each SVO, ensuring 
that it is in the present tense (e.g., \textit{``gave''} $\rightarrow$ 
\textit{``give''}). This is to ensure that the system increases the likelihood 
of establishing a consistent semantic relationship between the 
original tweet and SVO representation. The \texttt{VerbPresent} function uses 
the \texttt{NodeBox}~\cite{nodebox2016nodebox} linguistics library, which bundles 
various grammatical libraries for increased accuracy. Finally, we repeat this 
process for each tweet in the input file of tweet objects, and then pass the 
output to the \texttt{Clustering} component.

\subsection{Clustering}

The \texttt{Clustering} component ensures that each of the SVOs input from the 
\texttt{DataCleaning} component are built into a suffix tree. Once constructed, the 
SVOs represented by the suffix tree are merged to form world views. The 
\texttt{Clustering} component is also responsible for executing our initial 
trust metric evaluation; calculating a level of trust which the system 
associates with each world view.

Firstly, the \texttt{Clustering} component passes each SVO to the\\ 
\texttt{SuffixTreeConstruction} module in order to seed the construction of a 
suffix tree. We build the suffix tree by
using the semantic similarity between each component of the SVO tuple, in effect, 
creating what we define as a \textit{Subject-Verb-Object Semantic Suffix Tree} 
(SVOSST). Simultaneously, we construct the suffix 
tree through an on-depth and on-breadth process based on Ukkonen's 
suffix links~\cite{ukkonen1995line}. Initially, the suffix tree root is created 
and the first SVO is taken from the stack. The system then iterates through 
each of the constituent parts of the SVO tuple (subject, verb, and object), 
traversing the tree in order to find an overlap between the current word's 
synonym ring (synset) and the synset at each node. This is implemented using 
the WordNet~\cite{nltk2016nltk} corpus from the \texttt{NLTK} module. 
The system utilises the \texttt{synsets} function to retrieve the synset for 
each word, before adding it to the suffix tree. The complexity of SVO algorithm is $O(n)$, where $n$ is the number of words. Therefore the algorithm is linear in the number of words that needs to be processed and we do not anticipate any scalability issues when tested in big datasets.

Next, the \texttt{MergingPhase} module produces the final clusters which 
form our world views. In this function, the system uses each of the base 
clusters that have been identified by the \texttt{SuffixTreeConstruction} 
function, and decides how to cluster each of them using STC. More specifically,
the function executes an 
implementation of the STC algorithm~\cite{ilic2014suffix}, that we introduce 
as SVOSSTC. Our system asserts the maximum 
granularity of the clusters presented in the world view. This is because 
it is possible to produce base clusters with both a generic (length one) 
label, and a specific (length three) label with identical document sets. 
For example, the \texttt{ClustSim} function facilitates the avoidance of 
producing the output \textit{``David kill John''} and \textit{``John''}, 
when all of the documents stored within the \textit{``John''} cluster 
are the same as \textit{``David kill John''}.

Finally, the \texttt{ClusterTrustworthiness} component, analyses each tweet within a 
world view in the context of trusted sources in order to establish 
a measure for corroboration. A cumulative total of trusted entities is then 
recorded for each cluster, and divided by the total number of trusted sources 
that have been defined. The result of this calculation is then combined with 
the original cluster cardinality, excluding retweets, in order to give an 
indication of the overall trust of each cluster. This part of the \texttt{Clustering} 
component is highly extensible, providing an interface for various trust 
metrics to be applied to world views produced by the system.

\subsection{Visualisation}
\begin{figure}[ht!]
	\centering
	\frame{\includegraphics[width=.9\textwidth]{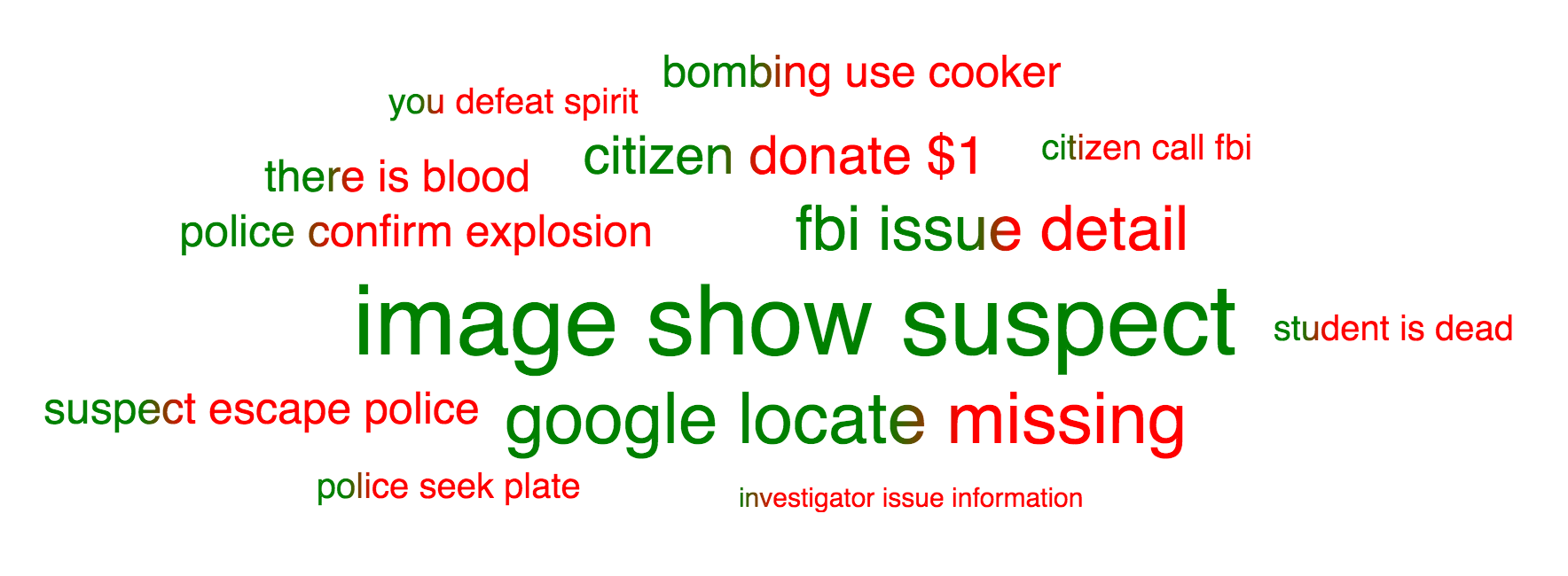}}
	\caption{Tag cloud produced by the \texttt{Visualisation} component}
	\label{fig:visualisation-design}
\end{figure}

The \texttt{Visualisation} component presents the information generated by the 
system to the end-user. This component forms the front-end of the system that is 
responsible for conveying our world views in the form of tag clouds. 
Figure~\ref{fig:visualisation-design} shows a tag cloud produced by the 
system based on a subset of real-world data from the 2013 Boston bombing. 
The tag cloud forms the main contribution of the \texttt{Visualisation} 
component.

The \texttt{Visualisation} component uses the score produced by the \\
\texttt{ClusterTrustworthiness} function in order to produce a coloured 
tag cloud. In the tag cloud, the size of the phrase is proportional to the 
cardinality of the cluster that the phrase represents, and the colour variation 
(between trustworthy in green, and misinformation in red) highlights possible 
misinformation identified. Therefore, the cluster \textit{``image show 
	suspect''} in Figure~\ref{fig:visualisation-design} shows a large 
cluster of information that contains only trustworthy sources corroborating 
each other.

\begin{figure}[ht!]
	\centering
	\frame{\includegraphics[width=.9\textwidth]{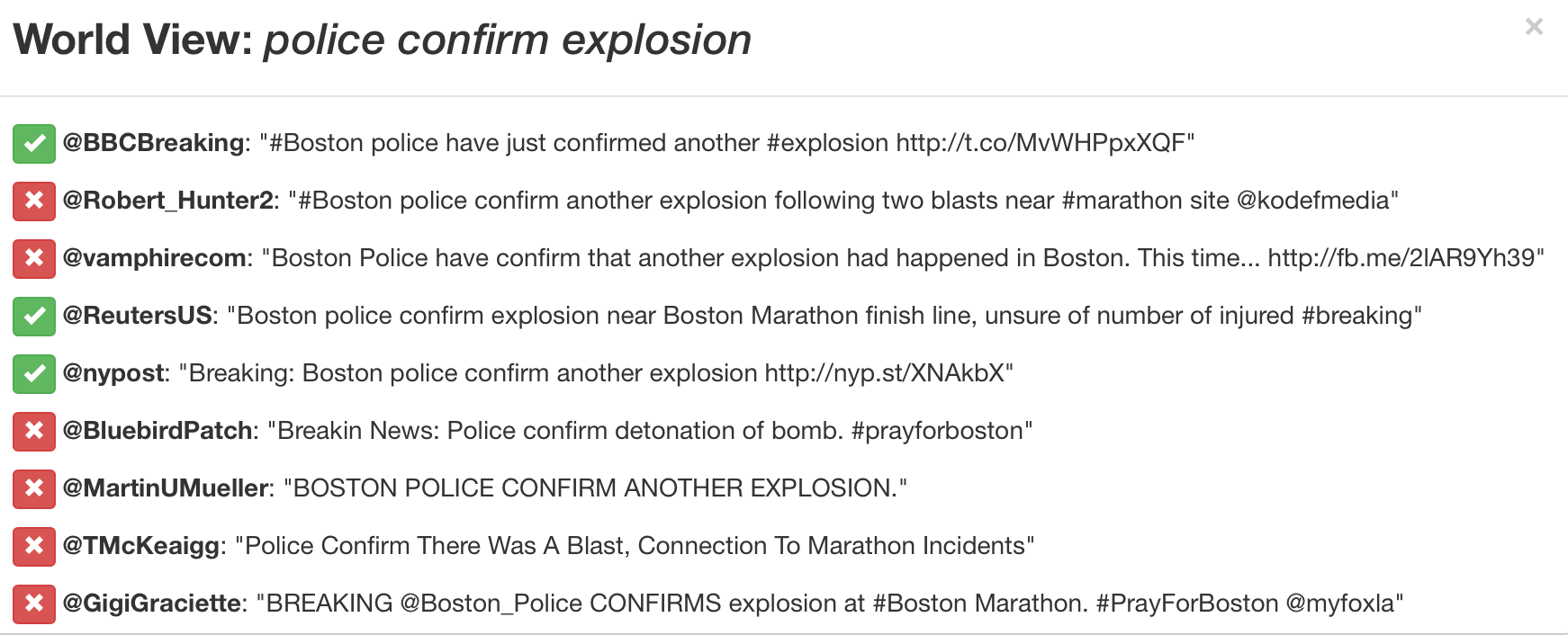}}
	\caption{World view alert produced by the \texttt{Visualisation} component}
	\label{fig:visualisation-alert}
\end{figure}

An in-depth overview of each world view is also generated in the 
\texttt{Visualisation} component when the corresponding element is 
selected from the tag cloud. The alert window (modal) we have developed 
is shown in Figure~\ref{fig:visualisation-alert}. This modal provides a 
detailed insight into the composition of the selected world view, and the 
trustworthiness score associated with each constituent element. 
Figure~\ref{fig:visualisation-alert}, presents the world 
view containing all of the tweets from the \textit{``police confirm 
	explosion''} tag shown in Figure~\ref{fig:visualisation-design}.

\section{Evaluation and Discussion}
\label{section:evadis}

\subsection{Method}

The main research challenge that we seek to tackle in this paper 
is the creation of a novel system that can understand datasets and determine 
consistent and corroborated information times. 
To evaluate our system in the context of this aim, and thereby assess its 
initial ability to allow an enhanced situational awareness in real-life 
scenarios, we used three datasets from a variety of crises 
that have occurred in recent years. In each of these crisis scenarios, Twitter 
was deemed to play a critical role in the efforts of emergency responders
attempting to understand the situation. The datasets used are:

\begin{itemize}
	\item Dataset 1 (\textbf{D1}): 172 randomly gathered tweets from 
	the 2016 Paddington train station incident, caused by an 
	individual standing on a bridge above the railway 
	line~\cite{gutteridge2016police};
	\item Dataset 2 (\textbf{D2}): 255 randomly gathered tweets from 
	the 2013 Boston bombing, where two pressure cooker bombs exploded 
	killing innocent civilians~\cite{eligon2013blasts}; and
	\item Dataset 3 (\textbf{D3}): 584 randomly gathered tweets from 
	the 2016 Ivory Coast beach resort attack, where a number people 
	were killed by a gun attack~\cite{aboa2016al}.
\end{itemize}

At this stage in our research, our aim was to assess the effectiveness of our
proposed system and also to incorporate the judgement of a human agent in
the assessment. Considering this point, we decided to use smaller dataset 
samples instead of emphasising large datasets, which would transfer the
focus to scalability.  Moreover, we do not seek to evaluate the trust component 
of the proposed system, noting our defined research challenge. These are areas
that will form central components of our future work in the space.

In summary therefore, \fixmetext{five} existing systems that use a variety of approaches to cluster 
information will be applied to \textbf{D1--D3}, in order to 
determine the top three world views produced (based on the 
cardinality of the cluster).

As mentioned in Section~\ref{section:litrev}, methods which required manual classification of tweets or manualy crafted lists with relevant words for events, such as the SVM approach proposed by Sakaki et al.~\cite{sakaki2010earthquake}, were not considered relevant for comparison to our work. Our decision was informed by the fact that the system we propose in this paper does not assume prior knowledge of events. In a similar vein, classifiers which aim to identify assertions, questions and expressions based on an initial tweet, such as the one propoposed by Vosoughi~\cite{vosoughi2015automatic} are relevant for establishing tweets' stance but do not cluster tweets. Therefore, we decided to benchmark our system with approaches that utilise \fixmetext{STC, $k$-means and Latent Dirichlet allocation (LDA)} algorithms and whose code was publicly accessible. Our main criteria for selecting \fixmetext{$k$-means and LDA} clustering with cosine similarity, Carrot2 and Carrot2 STC are the fact that these algorithms do not require prior knowledge and that all three attempt to cluster tweets into different world views.

The systems used are:

\begin{itemize}
	\item System 1 (\textbf{S1}): $k$-means clustering with cosine 
	similarity;
	\item System 2 (\textbf{S2}): Carrot2 with 
	Lingo~\cite{osinski2004conceptual};
	\item System 3 (\textbf{S3}): Carrot2 with Suffix 
	Tree Clustering (STC);
	\item System 4 (\textbf{S4}): Human agent using manual clustering 
	techniques; and
	\fixmetext{ \item System 5 (\textbf{S5}): LDA clustering with cosine similarity.}
\end{itemize}

\fixmetext{The S1, S2, S3 and S5} clusters were selected 
as they represent the most important information likely to affect 
an individual's situation awareness. Furthermore, by reducing the 
scope of evaluation to three world views it is possible to analyse 
more influential (and critical) information in greater detail. 

The first approach we evaluate our system against is the $k$-means 
clustering algorithm (\textbf{S1}). \textbf{S1} is widely used 
in many document clustering 
approaches~\cite{ravindran2015k, steinbach2000comparison}. 
The algorithm performs iterative relocation to partition a dataset 
into clusters, locally minimising the distance between the cluster 
centres (centroids) and the data points (tweets represented as 
vectors). \textbf{S1} calculates the distance between each vector 
(tweet) using the cosine similarity measure~\cite{ozdikis2012semantic}), subsequently clustering the 
vectors (tweets) based on proximity to the centroid.

Carrot2 is an open-source clustering engine that automatically 
organises collections of documents into thematic categories. 
Lingo (\textbf{S2}) is the first of the Carrot2 algorithms 
that we evaluate against our approach. \textbf{S2} is notable 
as it reverses the traditional clustering pipeline by first 
identifying cluster labels and then assigning documents to 
the labels to form the final clusters. It achieves this using 
Latent Semantic Indexing (LSI), which analyses the 
relationship between documents by assuming that similar words 
will occur in similar pieces of text~\cite{dumais2004latent}. 
\textbf{S2} exploits concepts found in the document 
through LSI, rather than identifying literal terms that are 
syntactically identical. 

STC (\textbf{S3}) traverses a 
suffix tree in order to identify words that occur in more 
than one document, under the assumption that common topics 
are expressed using identical sequences of terms. Each of 
these words gives rise to a base cluster where the nodes 
contain information about the documents in which each 
phrase appears. The base clusters are then merged using 
a predefined threshold, retaining only those documents 
that meet a predefined minimal base cluster score.

The fourth approach used is a human agent (\textbf{S4}). In order to establish a fair comparison, we decided to recruit an individual that had experience with social media and could understand clustering techniques. Specifically, the person selected to cluster the tweets into different world views was a final year Trainee Solicitor studying for a Graduate Diploma in Law. Our decision was justified because we believed that this individual would use typical analytical reasoning to extract the world views. 

\fixmetext{The final approach we utlise is LDA, which assumes that documents represent mixtures of topics~\cite{blei2003latent}. Each topic is modeled as a probability distribution over a set of words. Thus, given a fixed number of topics (denoted by $n$), LDA creates a ``probabilistic graphical model''~\cite{blei2003latent} that estimates the probability of a topic's relevance to a particular document (i.e., $p(topic | document)$) and the probability of a word appearing given a particular topic (i.e., $p(word | topic)$). This allows LDA to compute the probability that a particular topic is relevant to a given document. We use the LDA implementation in the Python library gensim~\footnote{https://pypi.org/project/gensim/}; cosine similarity measure~\cite{ozdikis2012semantic} is used to determine to ensure reproducibility and consistency, we used the same random seed when performing LDA on each dataset.}

Our evaluation considers an objective human agent as an 
important benchmark in order to understand how the problem 
of clustering may be approached without the use of complex 
clustering systems. In order to conduct this evaluation, 
\textbf{S4} manually identifies a number of concepts 
present in the datasets. In the scenario where more than 
three concepts are identified, \textbf{S4} then 
reprocesses the candidate clusters in order to propose 
the top three clusterings (based on the cardinality of 
each set). We envisage that this approach will 
give us high accuracy for identifying the ground truth in 
the evaluation datasets. 

Before \textbf{D1--D3} can be evaluated \fixmetext{by \textbf{S1--S5}}, 
there are a variety of data formats that must be produced 
from the original data. For $k$-means clustering for
instance, we had to implement an approach to translate 
\textbf{D1--D3} to vectors of weighted word frequencies. \fixmetext{A similar process was performed for LDA algorithm}. Carrot2 on the other hand requires input documents to be in 
the Carrot2 XML format~\cite{carrot2016carrot}.

The evaluation will use \fixmetext{\textbf{S1--S5}} to 
analyse \textbf{D1--D3} to identify how well each system 
performs, including the one presented in this paper, and the quality of the world views created. 
\fixmetext{We also present a quantitative comparison of $k$-means, LDA and SVOSSTC where the focus is on the number of cluster created by these algorithms rather than the content and we discuss our results.}

\subsection{Results for qualitative analysis}

Table~\ref{tab:results-cosine}--Table~\ref{tab:results-LDA} 
\fixmetext{show the results of \textbf{S1--S5}} applied to 
\textbf{D1--D3}. A variety of 
results have been obtained, including keyword clusters 
(i.e., world views) that provide a high level overview of a 
situation, granular low-level clusters, and clusters 
that identify negation and semantic similarities.

\begin{table}[ht!]
	\footnotesize
	\centering
	\begin{tabularx}{.9\textwidth}{|p{1.9cm}|X|}
		\hline
		\textbf{} 
		& \textbf{S1}: \textit{$k$-means clustering with cosine similarity} 
		\\ \hline
		
		\textbf{D1}: \newline\textit{Paddington} 
		&	\textbf{C1.1.1}: Run, Unable, Follow \newline
		\textbf{C1.1.2}: Station, Tube, @DailyMirror \newline
		\textbf{C1.1.3}: Closed, Breaking, Jump, Threatening
		\\ \hline
		
		\textbf{D2}: \newline\textit{Boston} 
		&	\textbf{C1.2.1}: Continued, Crossed, Finish, Line \newline
		\textbf{C1.2.2}: Arrested, @BostonGlobe, Terror \newline
		\textbf{C1.2.3}: Eludes, Shuts, Hunt
		\\ \hline
		
		\textbf{D3}: \newline\textit{Ivory Coast} 
		& 	\textbf{C1.3.1}: Terrorist, @News\_Executive, Seaside \newline 
		\textbf{C1.3.2}: Guns, Machine, Gunmen, @DailyMirror \newline 
		\textbf{C1.3.3}: Witnesses, Way, @AFP
		\\ \hline	
	\end{tabularx}
	\linebreak 
	\caption{Results using $k$-means clustering with cosine similarity (\textbf{S1})}
	\label{tab:results-cosine}
\end{table}

\begin{table}[ht!]
	\footnotesize
	\centering
	\begin{tabularx}{.9\textwidth}{|p{1.9cm}|X|}
		\hline
		\textbf{} 
		& \textbf{S2}: \textit{Carrot2 with Lingo} 
		\\ \hline
		
		\textbf{D1}: \newline\textit{Paddington} 
		& 	\textbf{C2.1.1}: Paddington due to emergency services dealing \newline
		\textbf{C2.1.2}: Services are currently unable to run \newline
		\textbf{C2.1.3}: Service between Edgware Road and Hammersmith due
		\\ \hline
		
		\textbf{D2}: \newline\textit{Boston} 
		&	\textbf{C2.2.1}: News \newline
		\textbf{C2.2.2}: Released \newline
		\textbf{C2.2.3}: Blood
		\\ \hline
		
		\textbf{D3}: \newline\textit{Ivory Coast} 
		& 	\textbf{C2.3.1}: Shooting \newline 
		\textbf{C2.3.2}: Ivory Coast beach resort \newline 
		\textbf{C2.3.3}: Hotel in an Ivory Coast resort popular
		\\ \hline	
	\end{tabularx}
	\linebreak 
	\caption{Results using Carrot2 with Lingo (\textbf{S2})}
	\label{tab:results-lingo}
\end{table}

\begin{table}[ht!]
	\footnotesize
	\centering
	\begin{tabularx}{.9\textwidth}{|p{1.9cm}|X|}
		\hline
		\textbf{} 
		& \textbf{S3}: \textit{Carrot2 with STC} 
		\\ \hline
		
		\textbf{D1}: \newline\textit{Paddington} 
		& 	\textbf{C3.1.1}: Dealing with incident, Emergency services dealing, @GWRHelp \newline
		\textbf{C3.1.2}: Incident at Royal Oak, Police incident, Due to a police \newline
		\textbf{C3.1.3}: Station
		\\ \hline
		
		\textbf{D2}: \newline\textit{Boston} 
		& 	\textbf{C3.2.1}: Explosion, Boston Marathon \newline 
		\textbf{C3.2.2}: Bombing \newline
		\textbf{C3.2.3}: Victims, Blood, Run
		\\ \hline
		
		\textbf{D3}: \newline\textit{Ivory Coast} 
		& 	\textbf{C3.3.1}: Breaking, Shooting \newline
		\textbf{C3.3.2}: Beach \newline
		\textbf{C3.3.3}: Reports, Beach resort, Resort in Ivory Coast
		\\ \hline	
	\end{tabularx}
	\linebreak 
	\caption{Results using Carrot2 with STC (\textbf{S3})}
	\label{tab:results-stc}
\end{table}

\begin{table}[ht!]
	\footnotesize
	\centering
	\begin{tabularx}{.9\textwidth}{|p{1.9cm}|X|}
		\hline
		\textbf{} 
		& \textbf{S4}: \textit{Human} 
		\\ \hline
		
		\textbf{D1}: \newline\textit{Paddington} 
		& 	\textbf{C4.1.1}: No train service \newline
		\textbf{C4.1.2}: Man jumping off a bridge \newline
		\textbf{C4.1.3}: Several delays
		\\ \hline
		
		\textbf{D2}: \newline\textit{Boston} 
		& 	\textbf{C4.2.1}: Marathon explosion \newline
		\textbf{C4.2.2}: People give blood \newline
		\textbf{C4.2.3}: Images released showing suspect 
		\\ \hline
		
		\textbf{D3}: \newline\textit{Ivory Coast} 
		& 	\textbf{C4.3.1}: Terrorist attack on beach hotel \newline
		\textbf{C4.3.2}: Gunmen armed with machine guns \newline
		\textbf{C4.3.3}: Lots killed
		\\ \hline	
	\end{tabularx}
	\linebreak 
	\caption{Results using a human agent (\textbf{S4})}
	\label{tab:results-human}
\end{table}

\begin{table}[ht!]
	\footnotesize
	\centering
	\begin{tabularx}{.9\textwidth}{|p{1.9cm}|X|}
		\hline
		\textbf{} 
		& \textbf{S5}: \textit{LDA clustering} 
		\\ \hline
		
		\textbf{D1}: \newline\textit{Paddington} 
		&	\textbf{C1.1.1}: services, an, with, emergency, dealing, incident, not, running \newline
		\textbf{C1.1.2}: the, a, royal, at, oak, on, are, police \newline
		\textbf{C1.1.3}: to, royal, police, at, and, a, oak, due
		\\ \hline
		
		\textbf{D2}: \newline\textit{Boston} 
		&	\textbf{C1.2.1}: boston, to, the, marathon, of, in, suspect, bombing \newline
		\textbf{C1.2.2}: to, in, bombing, marathon, hospital, suspect, is, boston \newline
		\textbf{C1.2.3}: boston, marathon, suspect, explosion, the, bombing, at, police
		\\ \hline
		
		\textbf{D3}: \newline\textit{Ivory Coast} 
		& 	\textbf{C1.3.1}: in, beach, resort \newline 
		\textbf{C1.3.2}: in, hotel, ivory \newline 
		\textbf{C1.3.3}: in, resort, ivory
		\\ \hline	
	\end{tabularx}
	\linebreak 
	\caption{Results using LDA clustering (\textbf{S5})}
	\label{tab:results-LDA}
\end{table}

In what follows, we discuss the qualitative results of the 
evaluation in order to analyse the success of \textit{Subject-Verb-Object Semantic Suffix Tree Clustering}
(\textbf{SVOSSTC}) at identifying world views.

\clearpage
\subsection{Discussion on qualitative analysis}
Previous research, along with our findings, suggest that 
there is still a significant need to utilise a variety of 
techniques to improve situation awareness. We believe that 
our approach (\textbf{SVOSSTC}) makes a significant step 
forward in achieving this goal by providing a foundation for 
information credibility, and through blending syntax and 
semantics in the context of document clustering.

It was crucial to our evaluation that the systems \fixmetext{
(\textbf{S1--S5})} overcome the limitations discussed in 
Section~\ref{section:litrev}, by using the datasets 
(\textbf{D1--D3}) despite the information source, 
composition, and quantity. Through conducting the 
evaluation, there were four key themes that we identified, 
which affected the level of understanding an individual 
can obtain in a situation: (i) the level of granularity 
that each clustering method provides; (ii) the way in 
which negation is dealt with; (iii) the way in which 
syntax and semantics are used; and (iv) the impact of 
large quantities of data. The remainder of this section 
will discuss these themes.

\subsubsection{Granularity of Clusters}

Handling granularity is of crucial importance for delivering appropriate 
information to users, enabling them to match their 
information needs as accurately as 
possible~\cite{pfennigschmidt2009handling}. The need for 
granularity is enhanced when a variety of information 
sources are being combined into clusters, in order to 
avoid oversimplification of key points of information. 
However, some systems struggle with issues of generalisation, 
which \textbf{SVOSSTC} attempts to handle
more effectively in the context of crisis management.

In \textbf{S1}, it is evident that the system  
oversimplifies each incident (as shown in 
Table~\ref{tab:results-cosine}) by specifying a variety 
of keywords for each cluster label. As \textbf{S1} uses 
individual keywords for its clustering methodology, 
as opposed to phrases used with \textbf{SVOSSTC}, it 
is difficult to construct descriptive clusters whose 
labels have semantic dependency upon one another. 
Instead, little context is provided as to the situation 
described by \textbf{D1--D3}. Most notably, when 
\textbf{S1} clusters the Paddington dataset (\textbf{D1}) 
to produce the cluster \textit{``Closed, Breaking, Jump, 
	Threatening''} (\textbf{C1.1.3}), the only context we 
have about the root cause of the situation (where a 
man is jumping off a bridge) is the singular word 
\textit{``Jump''}.

In contrast, when \textbf{C1.1.3} is compared with the 
cluster \textit{``Man jump bridge''} (\textbf{CF.1.2}) 
produced by \textbf{SVOSSTC}, it is possible to 
clearly identify the difference in granularity. 
\textbf{SVOSSTC} demonstrates the 
ability to identify crucial contextual information 
within \textbf{D1}, thus outperforming the basic 
approach taken in \textbf{S1}. In the context of 
supporting an individual's situation awareness, this 
world view exemplifies how important granularity is.

The issue of granularity is further exemplified in \textbf{S3} which creates the 
most general labels, 
most notably \textit{``Dealing with incident''} 
(\textbf{C3.1.1}) and \textit{``Incident at Royal Oak''} 
(\textbf{C3.1.2}). Both examples restate the 
overarching theme of \textbf{D1}, but do not provide any additional contextual 
information. One reason for this is
that \textbf{S3}'s thresholds for clustering are crucial in the process of cluster formation. These 
thresholds are inherently difficult to tune, which results 
in variable results on some datasets. Furthermore, STC's phrase 
pruning heuristic tends to remove longer high-quality phrases, 
leaving only the less informative and shorter 
ones~\cite{weiss2001clustering}. We believe that through using 
the SVO representation it is possible to create world views 
that contain more structured and contextual information with 
\textbf{SVOSSTC}. This is supported by \textbf{CF.1.1--CF.1.3}.

Both \textbf{S2} and \textbf{S3} experience issues with granularity when 
analysing \textbf{D2}. Firstly, 
\textbf{S2} produces clusters which contain single keyword 
cluster labels \textit{``News''}, \textit{``Released''}, and 
\textit{``Blood''} (\textbf{C2.2.1--C2.2.3} respectively). 
Interestingly, \textbf{S2} is unable to capture the 
relationship between the components of certain tweets. 
In particular, that it was the \textit{``News''} being 
\textit{``Released''}. This is also an issue present in 
\textbf{S1}, where relationships amongst words are not 
considered. In contrast, \textbf{SVOSSTC} was able to 
produce \textit{``Police confirm explosion''} (\textbf{CF.2.3}), 
which is in fact the news that is being released by the 
police within \textbf{D2}.

Furthermore, when \textbf{S3} is applied to \textbf{D2}, 
it produces high level keyword descriptors for the resulting 
clusters. For example, \textit{``Victims, Blood, Run''} 
(\textbf{C3.2.3}) is comparable to the often poor clusters 
produced by \textbf{S1}. This keyword style of cluster 
labelling is frequent in \textbf{S3}, due to the lack of semantic 
similarity between keywords. In \textbf{SVOSSTC}, by using 
the SVO representation, we believe it is possible to identify 
semantic similarity in a structured framework. \textbf{SVOSSTC} ensures that all possible tweets are added to the target cluster if they have semantic equivalence. 
This is discussed further in Section~\ref{ch:5-semantics}.

\begin{table}[ht!]
	\footnotesize
	\centering
	\begin{tabularx}{.9\textwidth}{|p{1.9cm}|X|}
		\hline
		\textbf{} 
		& \textbf{SVOSSTC}: \textit{Our approach} 
		\\ \hline
		
		\textbf{D1}: \newline\textit{Paddington} 
		& 	\textbf{CF.1.1}: Service !run Paddington \newline
		\textbf{CF.1.2}: Man jump bridge \newline
		\textbf{CF.1.3}: Police storm platform
		\\ \hline
		
		\textbf{D2}: \newline\textit{Boston} 
		& 	\textbf{CF.2.1}: Image show suspect \newline
		\textbf{CF.2.2}: Google locate missing \newline
		\textbf{CF.2.3}: Police confirm explosion
		\\ \hline
		
		\textbf{D3}: \newline\textit{Ivory Coast} 
		& 	\textbf{CF.3.1}: Gunman attack resort \newline
		\textbf{CF.3.2}: Army evacuate beach \newline
		\textbf{CF.3.3}: Gunfire leave dead
		\\ \hline	
	\end{tabularx}
	\linebreak 
	\caption{Results using our approach (\textbf{SVOSSTC})}
	\label{tab:results-gsstc}
\end{table}

\fixmetext{Similar conclusions can be drawn for LDA clustering. Table~\ref{tab:results-LDA} reports the most probable words when running LDA with $n = 3$ on each individual dataset of tweets. For some topics – such as Paddington and Boston – LDA successfully isolates words that indicate different world views; for example, when run on the Paddington dataset, two topics mention the ``\textit{royal police}'' and another includes ``\textit{services not running}.'' However, the bag-of-words model inhibits the readability of LDA results, and, when compared to Table~\ref{tab:results-human} or Table~\ref{tab:results-gsstc}, LDA outputs qualitatively inferior summaries that do not represent a coherent set of non-overlapping world views. For example, a human agent may report ``\textit{No train service}'', and SVOSSTC yields ``\textit{Service !run Paddington}'', which is significantly clearer than a bag of words document that includes ``\textit{services}'' and ``\textit{not running}''.
}
The findings of the evaluation, with respect to the level 
of granularity, demonstrate the ability for \textbf{SVOSSTC} 
to produce world views that outperform \textbf{S1}, 
\textbf{S2}, and \textbf{S3}.

\subsubsection{Handling Negation}\label{ch:5-negation}

The role of negation has been acknowledged as important in the 
field of linguistic analysis~\cite{wiegand2010survey}. In 
application domains such as sentiment analysis, this phenomenon 
has been widely studied and is considered to be crucial to the 
methodology~\cite{lapponi2012representing}. 
However, understanding and addressing the role of negation in 
clustering systems is often overlooked. As negation is a 
common linguistic structure that affects the polarity of 
a statement, it is vital to be taken 
into consideration when clustering information. Furthermore, 
in the context of situation awareness it is not possible 
to satisfy Endsley's three levels of situation awareness 
(perception, comprehension, and projection)~\cite{endsley1995toward} 
without negation. This is because to perceive an environment 
in the correct context, in order to comprehend the interpretation 
of these perceptions and subsequently project upon them, negation 
must be present to enable polarity to be considered.

There are numerous examples of unordered keywords representing 
cluster labels in the systems evaluated. In short, these 
systems ignore the basic concept of negation. \textbf{S1} 
using the Paddington dataset (\textbf{D1}) is a good example 
of this phenomenon. It produces the cluster 
\textit{``Run, Unable, Follow''} (\textbf{C1.1.1}) highlighting 
how unordered keywords, especially those affecting phrase 
polarity, cannot be mixed together. Specifically, the decision 
is left to the end-user to gauge whether \textit{``Unable''} 
refers to \textit{``Run''} or \textit{``Follow''} (it is in 
fact the former). The issue with ignoring negation is further 
exemplified in \textbf{C1.1.1}, where the keyword 
\textit{``Run''} is included in the cluster label produced. 
However, in \textbf{D1} the incident at Paddington has 
prevented services from running, and therefore \textit{``!Run''} 
would be a far more representative cluster label (where 
\textit{``!''} indicates negation). \fixmetext{LDA clustering suffers similar fate due to the fact that the algorithm vectorises documents and uses bag-of-words functionality to produce clusters. Therefore, in traditional document 
clustering methods, such as \textbf{S1} and \textbf{S5}, the 
fact that the words may be semantically related and temporally 
related is not taken into account.}

Another issue present in 
\textbf{S1}, \textbf{S5} and \textbf{S2}, 
is that of extensive stop word removal. Words affecting phrase 
polarity can often be removed in these approaches,
affecting the overall quality of world views produced. When 
observing the performance of \textbf{S2} 
using \textbf{D1}, the cluster \textit{``Service between Edgware 
	Road and Hammersmith due''} (\textbf{C2.1.3}) suggests that 
there is indeed a service running. This is a false statement and serves as an example of how stop word removal 
affects polarity.

In contrast, a human agent (\textbf{S4}) deals with negation with 
success, as it is simple for a human to acknowledge latent phrase 
structure. Our approach (\textbf{SVOSSTC}) is also able to handle 
negation with similar levels of success to that of \textbf{S4}. 
In this sense, \textbf{SVOSSTC} produces the cluster 
\textit{``Service !run Paddington''} (\textbf{CF.1.1}) which 
fully encompasses this phenomenon. \textbf{SVOSSTC} achieves 
this result by assigning negation to words identified during 
the SVO construction phase. It is necessary to 
retain all candidate stop words in the input, as they are used 
to generate the correct part-of-speech tags for this SVO 
processing procedure. This is crucial in the context of crisis management, and could 
cause significant problems if not addressed by systems producing 
world views.

Interestingly, \textbf{S3} performs slightly 
better than other existing approaches when analysing \textbf{D1--D3}. 
However, this may be due to the variety of words in the dataset 
itself, as \textbf{S3} is unable to deal with semantically related 
negation. For example, if there were information such as 
\textit{``There is not a service running at Paddington''} and 
\textit{``There is no service running at Paddington''}, 
\textbf{S3} would be unlikely to yield the single cluster result 
(due to \textit{``not''} and \textit{``no''}) that is possible 
with \textbf{SVOSSTC}.

The evaluation, in the context of handling negation, clearly 
demonstrates the ability for \textbf{SVOSSTC} to perform at a 
similar level to \textbf{S4}, whilst outperforming \textbf{S1--S3} 
in the context of crisis management.

\subsubsection{Syntax and Semantics}\label{ch:5-semantics}

Traditional clustering algorithms do not consider the semantic 
relationship between words, which means they cannot accurately 
group documents (tweets) based on their meaning~\cite{wei2015semantic}. 
Thus clustering is based on syntax alone, 
which is not sufficient to cluster a large quantity of structurally 
inconsistent data from a variety of sources. To overcome these issues, 
it is critical that semantic reasoning is used when clustering 
tweets, improving the resultant world views when compared 
to classical methods.

\textbf{S2} and \textbf{S3} 
highlight some of the semantic issues when analysing the Ivory 
Coast dataset (\textbf{D3}). For example, \textbf{S2} produces 
the clusters \textit{``Ivory Coast beach resort''} (\textbf{C2.3.2}) 
and \textit{``Hotel in an Ivory Coast resort popular''} 
(\textbf{C2.3.3}) which contain exactly the same contextual 
data, and are highly similar with respect to semantics. 
Therefore, these concepts should be clustered into a single 
world view. However, \textbf{S2} is unable to produce the 
desired result as it attempts to identify certain dominating 
topics, called abstract concepts, present in the search. 
\textbf{S2} then picks only such frequent phrases that best 
match these topics. This means that recurring and semantically 
similar phrases may not be clustered.

Our approach (\textbf{SVOSSTC}) avoids such issues as highlighted 
in the two paragraphs above because it 
does not produce non-SVO cluster labels, and once SVO tuples have 
been generated it is possible to address semantic similarity in 
the constituent parts of the SVO representation. For example, 
\textbf{SVOSSTC} produces the cluster \textit{``Gunman attack 
	resort''} (\textbf{CF.3.1}), which we perceive to outperform a 
cluster label simply stating the location of the attack. 
Furthermore, the issue of semantics can also be seen in 
\textbf{S3} using \textbf{D3}. In this system, each of the 
top three clusters (\textbf{C3.3.1--C3.3.3}) effectively 
reduce to the single cluster \textbf{CF.3.1} produced by 
\textbf{SVOSSTC}. By extracting the semantic meaning of a 
phrase using SVO, it is possible to understand that 
\textit{``Shooting''} (\textbf{C3.3.1}), \textit{``Beach''} 
(\textbf{C3.3.2}), and \textit{``'Resort...'} (\textbf{C3.3.3}) 
are more easily represented in \textbf{CF.3.1}. 

A major limitation of \textbf{S3}, and STC more generally, 
is that if a document does not include the phrase which 
represents a candidate cluster, it will not be included 
within that cluster, despite the fact that it may still be 
relevant~\cite{zhang2004semantic}. This is corrected using 
\textbf{SVOSSTC}, which takes this into consideration through
the use of word nets (Section~\ref{section:step2}).

Another interesting observation arises with \textbf{S3} when 
using the Paddington dataset (\textbf{D1}). In this dataset, 
\textbf{S3} cannot identify the similarities between 
\textit{``Dealing with Incident, Emergency Services Dealing''} 
(\textbf{C3.1.1}) and \textit{``Incident at Royal Oak, Police 
	Incident, Due to a Police''} (\textbf{C3.1.2}), despite them 
relating to exactly the same concept. This exemplifies that 
syntactic structure is often too strong and does not allow 
flexibility in the linguistic style of posts 
in our evaluation datasets. Instead, the clustering methodology 
produces many overlapping and semantically similar clusters.

However, there still exist several challenges, such as 
polysemy, high dimensionality, and extraction of core 
semantics from texts, which \textbf{SVOSSTC} needs to 
address more effectively to fully exploit the value of 
social media data in the context of syntax and semantics.

\subsubsection{Information Overload}\label{ch:5-overload}

Large datasets can allow for a much deeper insight into a 
scenario by providing a comprehensive, in-depth overview. 
Despite this being useful in understanding situations and 
making decisions, when there is too much information to 
process, there can be issues with information 
overload~\cite{rodriguez2014quantifying}, making the 
task of document (tweet) clustering increasingly difficult. 
In order to support an individual's situation awareness, 
systems must allow end-users to fully comprehend 
large datasets, to ensure that key themes can be extracted
(this relates to our world views point).

The damaging effects of information overload are present 
in a human agent (\textbf{S4}) using the Ivory Coast 
dataset (\textbf{D3}), the largest of our evaluation 
datasets (584 tweets). \textbf{S4} is unable to identify 
critical concepts that our approach (\textbf{SVOSSTC}) 
was able to extract from \textbf{D3}, including 
\textit{``Army evacuate beach''} (\textbf{CF.3.2}). This 
is an important milestone for \textbf{SVOSSTC}, as 
\textbf{S4} is often considered to have a good understanding 
and perception of different world views in a situation. 
However, according to the principle of bounded rationality, 
humans will only explore a limited range of 
alternatives and will consider a subset of the decomposition 
principles in order to make the task cognitively 
manageable~\cite{simon1996sciences}. This highlights 
the need for approaches, such as \textbf{SVOSSTC}, that are 
able to identify world views, to alleviate the 
informational capacity suffered by \textbf{S4}. It is 
this capacity that often reduces \textbf{S4}'s effectiveness 
in crisis situations when compared to other systems.

In contrast, \textbf{S4} performs increasingly better 
than \textbf{S1--S3 and S5} when analysing \textbf{D1--D3}, as 
it produces more effective cluster labels. Interestingly, 
the output of \textbf{S4} is similar to \textbf{SVOSSTC} 
for all datasets, demonstrating the success of the system 
in performing at a near human level. This is opposed to 
\textbf{S1--S3 and S5} which often struggle to produce labels that 
come with large amounts of information at a granular level.

It is important to note that despite the success in 
evaluating \textbf{SVOSSTC} against \textbf{S4}, the 
findings represent the process considered by a single 
objective human agent. We do accept the argument that it may 
be better to use a panel of individuals/judges and use a consensus 
for this approach. Multiple individuals could remove any bias
or unforeseen issues and therefore this should be pursued as
an avenue of future work. Generally however, in the context of information 
overload in our evaluation, it does demonstrate some of 
the major successes of \textbf{SVOSSTC} which facilitate 
a close to human clustering methodology.

\fixmetext{
\subsection{Results for quantitative analysis}
For our quantitative analysis, we focus on how well cluster algorithms categorise tweets and compare these results to the clusters SVOSSTC generates. We therefore focus our analysis on k-means and LDA algorithms. Figure~\ref{fig:k-means3}--Figure~\ref{fig:SVOSSTC} 
show the results of \textbf{S1, S5 and SVOSSTC} applied to a single dataset consisting all
\textbf{D1--D3} tweets. A significant difference between the three algorithms is that \textbf{S1} and \textbf{S5} require a pre-determined number of clusters where our approach \textbf{SVOSSTC} does not. We run LDA and k-means with 3 clusters (the number of different datasets) and with 10 to allow for greater granularity, since datasets \textbf{D1--D3} comprise different world views (as demonstrated in our qualitative analysis) for the same event.

\begin{figure}[ht!]
	\centering
	\frame{\includegraphics[width=.75\textwidth]{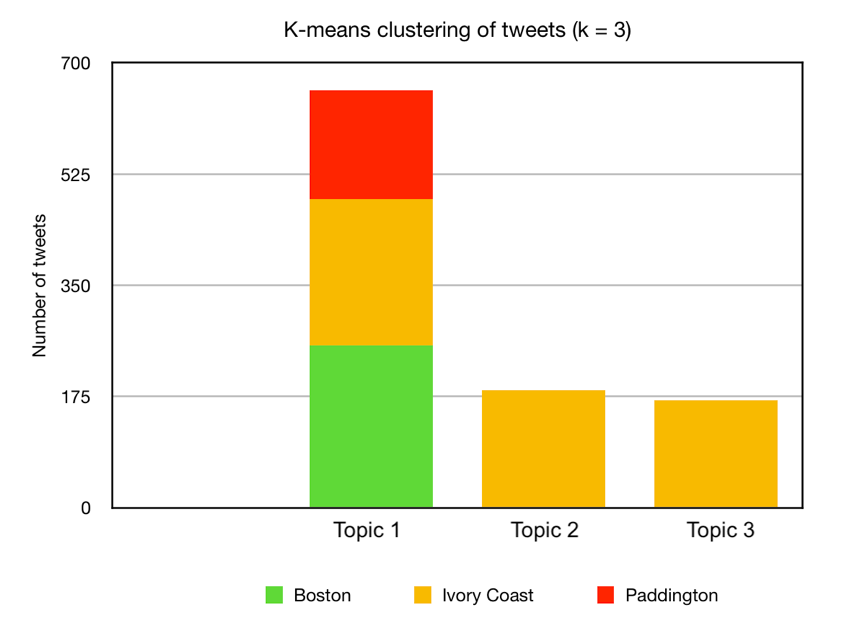}}
	\caption{Output from running k-means on combined dataset of tweets for three clusters, where topic 1 is Boston tweets, topic 2 is Ivory Coast tweets and topic 3 is Paddington tweets}
	\label{fig:k-means3}
\end{figure}

\begin{figure}[ht!]
	\centering
	\frame{\includegraphics[width=.75\textwidth]{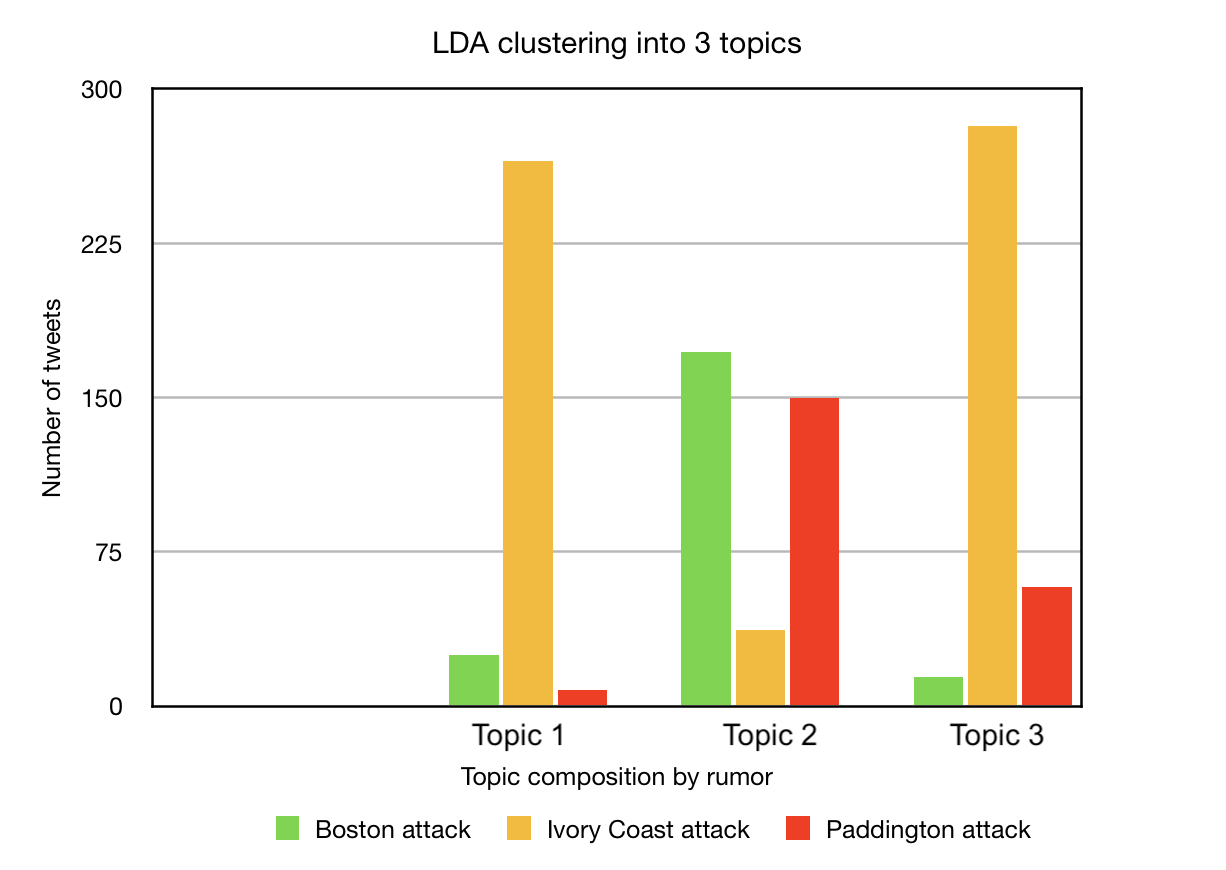}}
	\caption{Output from running LDA on combined dataset of tweets for three clusters, where topic 1 is Boston tweets, topic 2 is Ivory Coast tweets and topic 3 is Paddington tweets}
	\label{fig:LDA3}
\end{figure}

\begin{figure}[ht!]
	\centering
	\frame{\includegraphics[width=.75\textwidth]{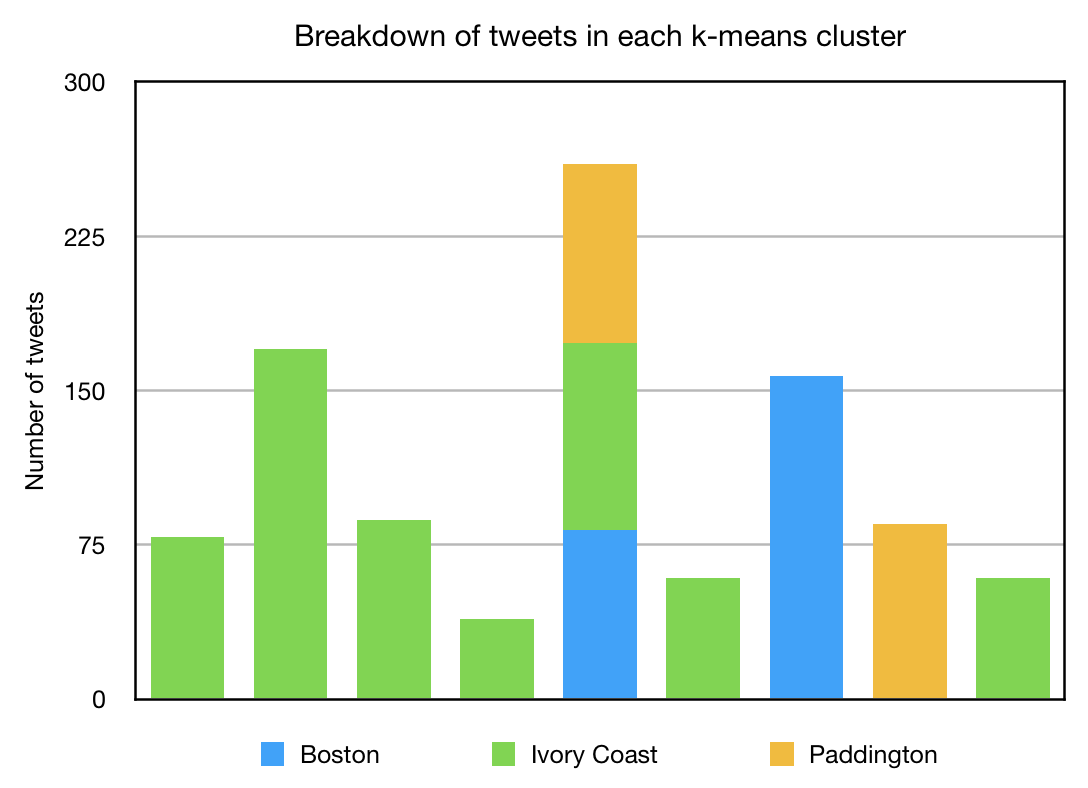}}
	\caption{Output from running $k$-means on combined dataset of tweets for ten clusters}
	\label{fig:k-means10}
\end{figure}

\begin{figure}[ht!]
	\centering
	\frame{\includegraphics[width=.75\textwidth]{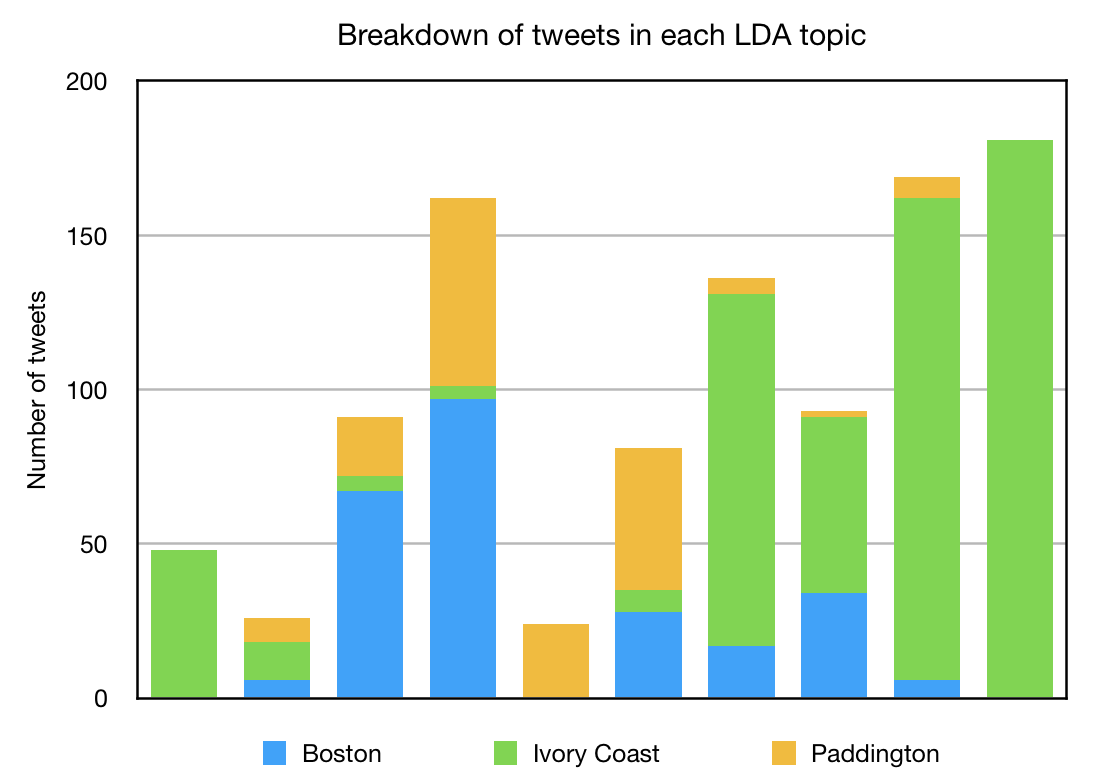}}
	\caption{Output from running LDA on combined dataset of tweets for ten clusters}
	\label{fig:LDA10}
\end{figure}

\begin{figure}[ht!]
	\centering
	\frame{\includegraphics[width=.75\textwidth]{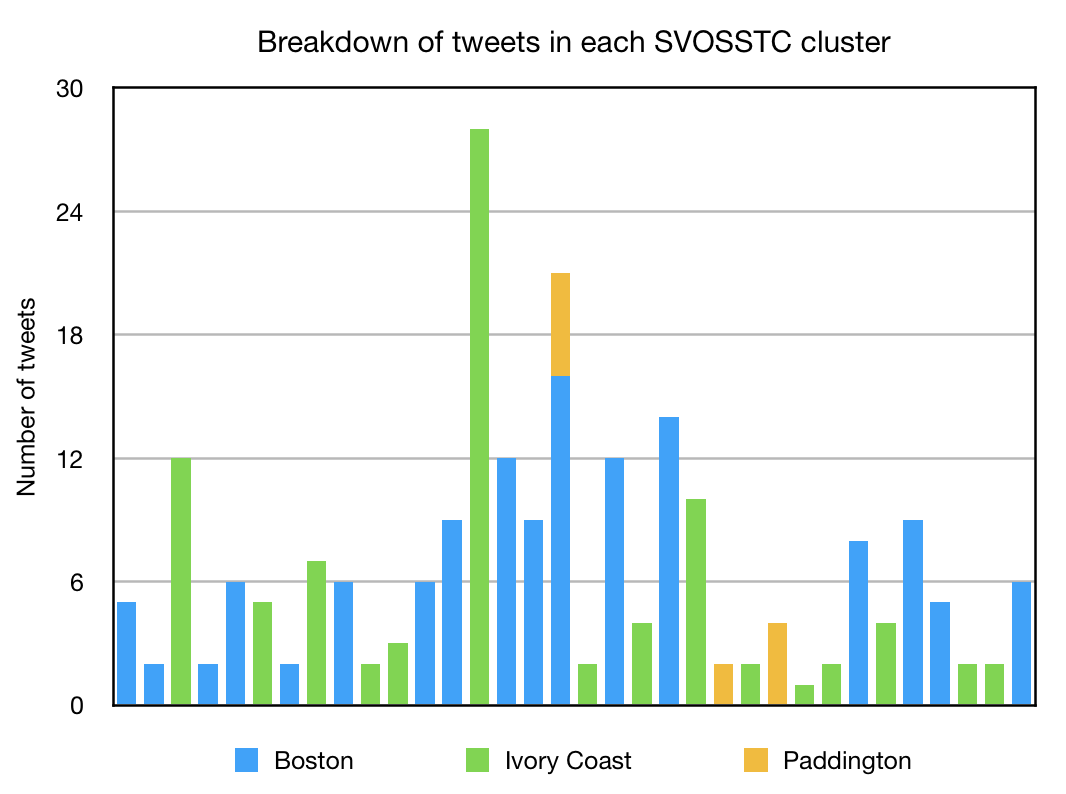}}
	\caption{Output from running SVOSSTC on combined dataset of tweets where number of clusters is not specified}
	\label{fig:SVOSSTC}
\end{figure}

In what follows, we discuss the quantitative results of the 
evaluation in order to analyse the success of \textit{Subject-Verb-Object Semantic Suffix Tree Clustering}
(\textbf{SVOSSTC}) at identifying clusters.

\clearpage 
\subsection{Discussion on quantitative analysis}
Our quantitative evaluation emphasises on the accuracy with which clustering algorithms assign tweets to clusters. We create an amalgamation of all the datasets by merging \textbf{D1-3} into a single dataset. Our aim is to identify whether the clusters created by algorithms are homogeneous (`pure'), thus all tweets derive from the same event (i.e., Boston attacks, Ivory Coast attacks or Paddington incident), or contain tweets from irrelevant events (heterogeneous). The presence of homogeneous clusters indicates that the algorithm is performing well. 

Figure~\ref{fig:k-means3} clearly demonstrates that k-means algorithm does not separately cluster tweets about the three different incidents. Although Clusters 2 and 3 consist only of tweets relating to the Ivory Coast, Cluster 1 contains all tweets on the Boston Marathon, all tweets on the Paddington incidents, and some tweets from the Ivory Coast attack. One potential explanation for these results is the sheer size of the Ivory Coast dataset; while  the first dataset contains 584 tweets regarding the Ivory Coast attack, the Boston dataset contains only 255 tweets and the Paddington only 172.  

The inclusion of retweets in the dataset suggests one potential reason for poor clustering behavior. In Figure~\ref{fig:k-means3}, 179 of the 185 tweets in Topic 2 are retweets; similarly, 146 of 169 tweets in Topic 3 are also retweets. As k-means does not perform any syntactic analysis, retweets --- which are extremely similar under cosine distance --- dominate Topics 2 and 3 and likely inhibit qualitative clustering. When increasing the number of clusters to 10, $k$-means performance improves, however a significant number of tweets from all three rumours is still miss-allocated, as Figure~\ref{fig:k-means10} illustrates.  

Unlike $k$-means, LDA does not definitively assign a particular document to a topic; instead, it produces a probability distribution representing the relevance of a document to all three topics. In Figures~\ref{fig:LDA3} and \ref{fig:LDA10}, we associate a given tweet with the topic LDA assigns the highest probability. Figure~\ref{fig:LDA3}  illustrates that LDA does not generate any homogeneous (i.e., high purity) cluster and all clusters contain tweets from all three incidents. Interestingly, when increasing the number of clusters to 10, there is no significant improvement in the performance, as Figure~\ref{fig:LDA10} demonstrates. 

When SVOSSTC is run on the combined dataset of tweets, 41 final clusters are generated. While some clusters associate tweets with their retweets, thus resembling $k$-means and LDA behaviour, others effectively congregate world views about a single incident, such as the presence of gunman in a resort in the Ivory Coast. However, SVOSSTC also yields one cluster that does not facilitate greater situational awareness. In this cluster many tweets discussing the Boston Marathon are clustered with others describing the Ivory Coast attack, probably because of the WordNet similarity of words used to describe the two terrorist attacks. We should note that SVOSSTC performs better than $k$-means and LDA, since in these algorithms there are clusters which contain tweets from all three incidents.} 

\fixmetextt{
	
\begin{table}[ht!]
	\footnotesize
	\centering
	\begin{tabular}{|c|c|}
		\hline
		& Average purity \\ \hline
		LDA     & 0.77           \\ \hline
		$k$-means & 0.94           \\ \hline
		SVOSSTC & 0.99           \\ \hline
	\end{tabular}
	\linebreak 
	\caption{Purity measures for LDA, $k$-means and SVOSSTC}
	\label{tab:results-purity}
\end{table}
	
An assessment of Figures~\ref{fig:k-means10}-\ref{fig:SVOSSTC} using the purity evaluation measure substantiates our qualitative assertion that SVOSSTC yields finer and more insightful clusters. LDA yields clusters with an average purity of 0.77 (see Table~\ref{tab:results-purity}), thus reflecting its inability to consistently separate tweets of a particular incident. Seven LDA clusters contain tweets from all three incidents. $K$-means reveals better performance with average purity 0.94. However, although most $k$-means clusters relate to a single incident, one includes dissimilar tweets from all three incidents, including train outages at Paddington, descriptions of the Boston bombing, and reports of gunmen in the Ivory Coast. 

SVOSSTC yields average purity 0.99; as described above, although one cluster contains tweets from two separate incidents, it groups tweets that pertain to similar terrorist attacks. Comparison of LDA, $k$-means, and SVOSSTC using the Rand index metric would likely also highlight the benefits of the semantic clustering approach. However, as LDA and $k$-means require choosing the number of topics, and SVOSSTC does not, purity was seen as a more natural performance metric for assessing SVOSSTC against $k$-means and LDA as it does not require computation of a confusion matrix with a fixed number of topics.}

\subsubsection{Summary}\label{ch:5-summary}

This evaluation focuses on the qualitative \fixmetext{ and quantitative} aspects of 
how our approach (\textbf{SVOSSTC}) performed in comparison 
to existing systems such as $k$-means clustering, Carrot2 
with Lingo, Carrot2 with STC,  a human agent and LDA. The 
findings were generally seen to support \textbf{SVOSSTC} 
as a useful, viable, and effective approach to addressing 
the issues of generating world views from social media data. 

In the discussion above, we 
analysed the overall success of our approach by identifying 
four key themes that came from the existing work. From this, 
a number of notable research contributions 
were recognised. Firstly, by using the SVO representation, 
we were able to establish a level of 
granularity that was beyond existing systems in the context 
of crisis management. Secondly, by handling negation as a 
core concept in our approach, this allowed us to maintain the 
polarity of tweets, despite the appearance of frequent 
phrases. Next, by utilising 
semantics within \textbf{SVOSSTC}, our approach facilitated 
the creation of meaningful clusters without a high level of unnecessary 
overlap. Finally, we were able to motivate the use of our 
system by highlighting the ability for it to overcome 
information overload in larger datasets. All of these problems often 
affect existing 
systems.

Focusing on the quantitative analysis, when applied to a dataset that amalgamates tweets from \textbf{D1--D3}, both k-means and LDA fail to reliably separate tweets into qualitatively useful categories that relate or distinguish the attacks, irrespective of the number of clusters. These results highlight advantages of the SVOSSTC approach. 

However, limitations were identified. In particular, the limitations are two-fold: 
the ability of the system to
extract an SVO representation from datasets regardless of the scenario; and 
the level to which semantic reasoning is an effective tool 
for overcoming polysemy and extracting core semantics from 
text. These points also raise questions regarding the 
refinements of the approach necessary, refinements which may
be possible as new and more advanced clustering techniques
become available and usable. \fixmetext{Furthermore, we intend to explore 
the use of Word2Vec~\cite{patterson2017deep} in the future enhancements of 
our approach as this may 
improve the functionality of topic extraction by addressing 
issues such as polysemy. Word2Vec is a computationally-efficient
set of models used to produce word embeddings, which have become popular
in the natural language processing field.} While we feel that the limitations
discovered are important issues, we do not believe that they
seriously undermine the contribution of this research and the
system proposed.  

Overall, from the results of the evaluation it is evident
that a majority of the findings were in support of 
\textbf{SVOSSTC} as an approach to facilitate the creation 
of world views to aid situation awareness.

\section{Conclusion and Future Work}
\label{section:con}
In this paper, we sought to support user and organisational 
situation awareness through the research and development of a
system to analyse social media data. The specific goal of our 
system was to facilitate the analysis of datasets of multiple 
posts, and allow the clustering of consistent posts into different 
world views. Having identified gaps in existing research, relating 
to the lack of semantic consideration and a syntactic approach 
that was too narrowly focused, we settled on a 
\textit{Subject-Verb-Object Semantic Suffix Tree Clustering} 
(SVOSSTC) approach in order to produce world views. The advantage 
of SVOSSTC was found in its ability to create semantically consistent 
clusters of information with succinct cluster labels, applying 
techniques observed in the Suffix Tree Clustering (STC) 
algorithm. 
Our evaluation supported this advantage as we discovered
that a majority of the findings were in support of SVOSSTC,
regardless of some caveats to its application.

There are various interesting options for future research. One area
is to explore alternative typologies such as 
Verb-Subject-Object (VSO) and Verb-Object-Subject 
(VOS) which are all present in Verb-Object (VO) languages 
such as English. Furthermore, the inclusion of other VO typologies
would enable better identification of data that can be encoded in a 
structured format (subject, verb, and object). This data can then be 
semantically clustered with minor alterations to the existing 
Semantic Suffix Tree Clustering (SSTC) implementation. 
This would allow for a more complete and consistent picture of a 
situation due to more representative data being included in the output 
of the system.

Another avenue of further consideration is the semantic 
relationship between words in the SVOSSTC algorithm.  A slight limitation of synonym rings (synsets) is the fact that the algorithm which links  synonymous words together may not contain an exhaustive list of words. Therefore in rare occasions, some words may not be recognised and can be omitted. Instead of 
focusing entirely on synsets as a measure of semantic 
similarity, other approaches can be utilised. For example, the use of lexical 
chains~\cite{barzilay1999using} could help to increase the 
accuracy of semantic clustering. This view is motivated by the 
characteristics of lexical chains being able to provide a context 
for the resolution of an ambiguous term, and being able to identify 
the concept that the term represents. For example, 
\textit{``Rome $\rightarrow$ capital $\rightarrow$ city''} represents 
a lexical chain. This demonstrates how concepts in world views 
could henceforth be semantically clustered in combination with 
the existing SVOSSTC approach, increasing the probability 
of finding a semantic overlap between words.

In our further work, we are also keen on experimenting with the
variety of new clustering approaches being published (e.g.,
\cite{liu2017compressed,kozlowski2017semantic}). 
These may increase the accuracy of our approach or the efficiency
of its use, and thereby add to its suitability as a system for 
better understanding real-world situations. While we did not
concentrate on the efficiency of the proposed approach (in general
or as compared to the other algorithms), this will be a key factor in our 
continued work given that ideally, our system will be used in
a `live' context such as an unfolding crisis scenario.
\fixmetext{Using this, we
expect to focus on large-scale datasets and conduct a range of 
head-to-head comparisons between the
various techniques proposed (and our own improved technique). This 
would allow us to better investigate its efficiency and performance at addressing
the key issues of clustering and world-view analysis.}

From a user focused perspective, another area which we could explore in
future work is how individuals use and respond to different systems 
(S1--S4) and the clusters they produce. In the ideal case, we would
also look to try this in a real-world event to gain as authentic a 
response from users as possible. This would provide further validation
of our proposed system and its utility. \fixmetext{Most importantly, 
as we look towards building on this research and addressing the issue 
of misinformation online, we will need to better understand how to dynamically
identify trustworthy sources in real-time. Whilst corroborating clusters
(world views) with predefined trusted sources is possible, it is not
infallible. Trustworthy sources will need to be relevant and updated
to consider the context of the scenario and the user of the system. 
In this way, our approach will be able to make significant progress
on addressing the misinformation problem online.}



\begin{backmatter}

\section*{Competing interests}
The authors declare that they have no competing interests.

\section*{Funding}
There are no funding bodies to acknowledge for this research.

\section*{Author's contributions}
This article is the product of research led by Charlie Kingston (CK), in collaboration with Jason R. C. Nurse (JN), Ioannis Agrafiotis (IA) and Andrew Burke Milich (AM). As a result, the core research and experimentation was conducted by CK, with JN and IA providing direction and guidance during the research; AM also contributed to the experimentation and comparison of the proposed approach. All parties assisted in the journal manuscript drafting stage. All authors read and approved the final manuscript.

\section*{Acknowledgements}
There are no acknowledgements at this time.

\section*{Availability of data and material}
The Twitter data used in this study is not shareable due to the company's terms of service.

\vspace{3em}


\bibliographystyle{bmc-mathphys} 
\bibliography{references}      





%
%

\clearpage




%

\end{backmatter}
\end{document}